\documentclass[fleqn,%
aip,  
amsmath,amssymb,
reprint,%
groupedaddress
]{revtex4-1}
\usepackage{amsfonts,amsbsy,amscd,amsgen,amsthm,dsfont,amsmath,amssymb,mathrsfs}
\usepackage[colorlinks=true,allcolors=blue]{hyperref}
\usepackage[sort&compress]{natbib}
\usepackage{lipsum}
\usepackage{amsfonts}
\usepackage{graphicx,subfigure}
\usepackage{epstopdf}
\usepackage{algorithmic}
\usepackage{placeins}

\usepackage{amsopn}
\usepackage[normalem]{ulem} %to strike the words

\graphicspath{{figures/}} 
\usepackage{bm} %italic vectors
\usepackage{multirow}
\theoremstyle{definition}
\usepackage{xcolor}
\usepackage{enumitem}
\usepackage[utf8]{inputenc}
\usepackage[T1]{fontenc}
\newcommand{\id}{\mathrm d}
\newcommand{\vc}{\mathbf}

\renewcommand{\tilde}{\widetilde}

\DeclareMathAlphabet\mathbfcal{OMS}{cmsy}{b}{n}

\newcommand{\R}{\mathbb{R}}

\newcommand{\grad}{\nabla}

\newcommand{\cb}{\color{black}} 
\newcommand{\win}{\mathbf{W_{\text{in}}}}
\newcommand{\wout}{\mathbf{W_{\text{out}}}}

\renewcommand{\Re}{\mbox{Re}}
\renewcommand{\Im}{\mbox{Im}}

\begin{document}
\title{Model-assisted deep learning of rare extreme events from partial observations}
\author{Anna Asch} 
\affiliation{Department of Mathematics, Cornell University, 310 Malott Hall, Ithaca, NY 14853}
\author{Ethan Brady}
\affiliation{Department of Mathematics, Purdue University, 150 N. University Street, West Lafayette, IN 47907}
\author{Hugo Gallardo} 
\affiliation{Department of Mechanical Engineering, The University of Texas Rio Grande Valley, 1201 W. University Drive, Edinburg, TX 78539}
\author{John Hood}
\affiliation{Department of Mathematics, Bowdoin College, 8600 College Station Brunswick, ME 04011}
\author{Bryan Chu}
\author{Mohammad Farazmand}
\email{farazmand@ncsu.edu}
\thanks{corresponding author}
\affiliation{Department of Mathematics, North Carolina State University, Raleigh, NC 27695-8205, USA}

\begin{abstract}
	To predict rare extreme events using deep neural networks, one encounters the so-called small data problem because even long-term observations often contain few extreme events. Here, we investigate a model-assisted framework where the training data is obtained from numerical simulations, as opposed to observations, with adequate samples from extreme events. However, to ensure the trained networks are applicable in practice, the training is not performed on the full simulation data; instead we only use a small subset of observable quantities which can be measured in practice. We investigate the feasibility of this model-assisted framework on three different dynamical systems (R\"ossler attractor, FitzHugh--Nagumo model, and a turbulent fluid flow) and three different deep neural network architectures (feedforward, long short-term memory, and reservoir computing). 
	In each case, we study the prediction accuracy, robustness to noise, reproducibility under repeated training, and sensitivity to the type of input data. In particular, we find long short-term memory networks to be most robust to noise and to yield relatively accurate predictions, while requiring minimal fine-tuning of the hyperparameters.
\end{abstract}

\maketitle

\begin{quotation}
Deep learning has proven largely effective in predicting chaotic dynamical systems. However, to predict rare extreme events, one is confronted with the so-called small data problem: even long-term observations lack sufficient extreme events for training purposes.
We introduce a model-assisted framework where the deep neural network is trained using long-term simulations with adequate sampling from the extreme event regime. However, to ensure the trained network can be used with observational data, input variables only contain partial information comprising the quantities that can be measured in experiments. We examine the feasibility of this model-assisted framework on three neural network architectures, each trained with partial observations from three different dynamical systems.
\end{quotation}

\section{Introduction}
\label{sec:Introduction}
Extreme events such as rogue waves, earthquakes, epileptic seizures, and stock market crashes are rare but have devastating humanitarian, environmental, and financial consequences. 
Real-time prediction of these events is essential for optimal response management and mitigation of their most adverse consequences~\cite{comfort2010,farazmand2019a,Shafiei18}. 

Although deep learning has proven largely effective in predicting chaotic systems, to predict extreme events, one is confronted with the \emph{small data problem}: even long-term observations contain relatively few extreme events.
As a result, the available observational data lacks enough samples from the extreme event regime for training purposes. 
In addition, the full state of the system is often inaccessible in practice. This is especially the case for spatiotemporal systems where the state can only be measured at sparse sensor locations~\cite{Callaham2019,Chu2021,Kramer2017}. Therefore, the training must be carried out using a limited set of system observables. 

\begin{figure*}
	\centering
	\includegraphics[width=0.8\textwidth]{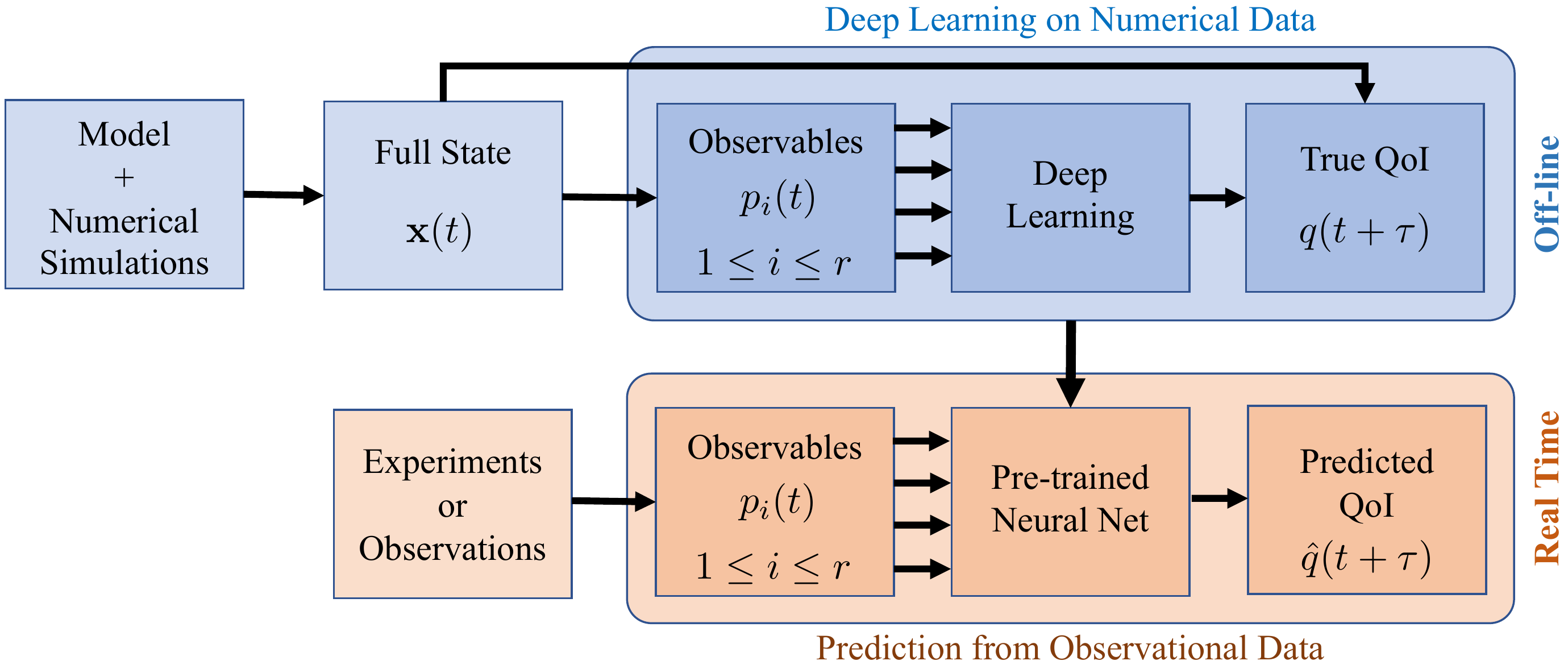}
	\caption{Schematic diagram of the program followed in this paper. We use a two-phase procedure to make predictions. In the offline phase, numerical simulations of a model are used to train a neural network. The trained neural network is then used in real-time to predict extreme events from observations.
	}
	\label{fig:schem}
\end{figure*}

Here we consider a model-assisted framework that addresses these sampling and partial observation issues. As depicted in figure~\ref{fig:schem}, the proposed framework has two components:
\begin{enumerate}
	\item Off-line: A computationally expensive component which is carried out off-line and only once. This step involves long-term numerical simulations and training a deep neural network.
	\item Real-time: A computationally inexpensive component that uses observational data as input for the pre-trained neural network to make real-time predictions.
\end{enumerate}
In the off-line step, we use a mathematical model of the system to generate long-term
simulations which contain enough extreme event samples to train a deep neural network. Although the simulation results contain the full state of the system $\vc x(t)$, we do not use this complete information for training the neural network.
Instead, we use this information to generate time series of the observables $p_i(t)$, i.e.
a set of quantities which can be actually measured in practice.
The observable time series are then used to train a deep neural network. 
The trained network uses the observations $p_i(t)$ as input to predict the future values of a quantity of interest $q(t+\tau)$, where $\tau$ denotes the prediction time. 
The quantity of interest (QoI) refers to a scalar quantity which is relevant to extreme events. For instance, in the case of ocean rogue waves, QoI is the maximum surface height~\cite{Farazmand2017b}, or in climatology, QoI may refer to the 
mean surface temperature~\cite{mendez2020}.

Once trained, the deep neural network is used in the real-time component to make predictions using the available observations $p_i(t)$. This component only relies on observational data and is quite fast since it uses a pre-trained network. We emphasize that the off-line component is designed to address the 
practical issues faced in predicting extreme events; mainly that in practice not all degrees of freedom can be measured (incomplete observations). 
Furthermore, the long-term numerical simulations provide enough sampling from 
extreme events to overcome the small data problem. Although we do not do so here, one can use rare event sampling techniques to reduce the amount of numerical simulations required~\cite{bucklew2004,Dematteis2018,Mohamad2018}. 

We examine the feasibility of this framework (see figure~\ref{fig:schem}) for predicting extreme events by considering three different dynamical systems and
three different neural network architectures: feedforward (FF), long short-term memory (LSTM) and reservoir computing (RC). For different architectures, we study their accuracy, robustness to noise, sensitivity to hyperparameters, reproducibility under repeated training, and dependence on the input data. 

\subsection{Related work}
As applications of machine learning have proliferated, they have been extensively used in predicting, analyzing, and discovering chaotic dynamical systems.
We refer to Refs.~\cite{brunton2020, fawaz2019, karniadakis2021, brenner2021,  Pathak2017} for recent reviews of applications of machine learning methods in dynamical systems. Here, we only review studies that use machine learning in the context of extreme events.

% work that uses full system information

Ding et al.~\cite{ding2019} design a hybrid recurrent neural network structure for extreme event prediction that addresses the scarcity of extreme events by introducing a specialized loss function inspired by extreme value theory.
Qi and Majda~\cite{majda2020} use a relative entropy loss function together with a convolutional neural network to predict extreme events in a truncated Korteweg–de Vries equation {\cb (also see Rudy and Sapsis~\cite{Rudy2021})}.
Senthilvelan et al.~\cite{senthilvelan2021} use standard deep learning methods to predict extreme events
in a parametrically driven mechanical system (see also Refs.~\onlinecite{lellep2020, narhi2018, yeditha2020}).

% work that uses reduced-order modeling
The above studies rely on the complete system state for their prediction tasks. However, such complete information is often unavailable in practice.
One remedy is to use reduced-order models that only rely on a subset of the state variables.
For instance, Wan et al.~\cite{wan2018} demonstrate that incorporating physical principles and reduced-order modeling in the neural network architecture can improve predictive accuracy and interpretability. 
Although reduced-order methods involve less information than the full system state, they still require solving a model which is not ideal for fast, real-time predictions. 

% work that uses partial system information
There are relatively few studies that examine the capacity of deep learning for predicting extreme events based only on observational data.
Chattopadhyay et al.~\cite{hassanzadeh2020} make model-free prediction of extreme-causing weather patterns using large-scale circulation data
and surface temperature to train a capsule neural network (CapsNet). They find that supplementing the circulation data with surface temperature significantly enhances 
the prediction accuracy. {\cb Rudy and Sapsis~\cite{Rudy2021b} use an LSTM network to predict intermittent aerodynamic fluctuations from discrete pressure measurements on an airfoil.}
Chattopadhyay et al.~\cite{subramanian2019} quantify the statics of a Lorenz 96 model using only the slow variables of the system as input data.
They find that LSTM and RC networks predict the heavy-tailed statistics (i.e., rare events) reasonably well, while FF networks fail to capture the tail.
Pyragas and Pyragas~\cite{pyragas2020} use RC networks to predict and mitigate extreme events in a FitzHugh-Nagumo system from a single quantity containing global information.

Here, to mimic real applications, we also assume that only a small number of observable time series are available as network inputs. 
{\cb Unlike previous studies that often consider one dynamical system or one network architecture, we conduct a comprehensive study by applying three different neural network architectures to three different dynamical systems which exhibit extreme events. In particular, we seek to address the following questions:
1. Accuracy: Is one neural network architecture more skillful in predicting extreme events across different dynamical systems? 2. Sensitivity to noise: Is a particular architecture more robust to observational noise? Furthermore, does training on clean simulation data outperform training on noisy data. 3. Sensitivity to hyperparameters: Which network requires less  hyperparameter fine tuning. 4. Reproducibility: Are the results reproducible under retraining? 5. Sensitivity to input data: Do the networks perform equally well when trained on different types of input data?}

\subsection{Outline of the paper}
This paper is organized in the following manner.
In section~\ref{math_prelim}, we introduce the set-up and notation, and outline the dynamical systems studied here. Section~\ref{MLarch} details the three deep learning structures we use. In section~\ref{results}, we present our numerical results. Section~\ref{conclusion} contains our concluding remarks. 

\section{Preliminaries and set-up} \label{math_prelim}
In this section, we introduce the dynamical systems set-up for extreme events.
In particular, we discuss three systems which are used later in this paper to demonstrate our results.

\subsection{Dynamical systems set-up} 
\label{dsnotes}
We consider autonomous dynamical systems defined by a set of ordinary differential equations (ODEs),
\begin{equation}\label{ode1}
	\dot{\mathbf{x}} = \mathbf{F}\left(\mathbf{x}\right),\quad
	\mathbf{x}(0) = \mathbf{x}_0,
\end{equation}
where the state $\vc x(t) = (x_1(t), x_2(t), \ldots, x_n(t))$ belongs to $\mathbb{R}^n$ for all $t\geq 0$. 
The vector field $\vc F : \R^n\to \R^n $ is a potentially nonlinear map. 
ODE~\eqref{ode1} may model a finite-dimensional system 
or arise from a finite-dimensional approximation of a partial differential equation (PDE), as is common in numerical discretization of PDEs. 
We denote the solution map of the system by $S^t:\mathbb R^n\to \mathbb R^n$ which maps an initial condition $\vc x_0$
to its time-$t$ state $\vc x(t)$.

We consider a scalar quantity of interest $q: \R^n \to \R$ whose evolution is related to extreme events. This quantity is problem dependent; 
for instance, for rogue waves, the quantity of interest is the maximum wave height~\cite{Farazmand2017b} while, in turbulence, it may be energy dissipation~\cite{PRE2016,blonigan2019}.
Evaluated along a trajectory $\vc x(t)$ of the system, the quantity of interest generates the time series $q(t) := q(\vc x(t))$. 
An extreme event refers to an episode where the quantity of interest is unusually large. More specifically, 
we say an extreme event has taken place if the quantity of interest $q(t)$ exceeds a prescribed threshold $q_e$. 
The case where extreme events correspond to unusually small values of QoI can be handled similarly by redefining the quantity as $q\mapsto -q$.

We seek to predict extreme events before they take place. 
For a prediction time $\tau$, we use the available information up to the present time $t$ to predict the future value of the QoI $q(t+\tau)$. 
Of course, if the full state $\vc x(t)$ is accessible, the ODE can be integrated numerically to estimate
$q(t+\tau) = q \left( S^\tau \left(\vc x(t)\right) \right)$.

However, in practice, we often lack complete information about the full system state $\mathbf{x}(t)$. 
Instead, we assume that only \textit{partial observations} $\vc p = (p_1,p_2,\cdots,p_r) \in \R^r$, containing $r$ unique measurements, are available where $r\ll n$.
The $i$-th component of the observations is given by the map $p_i:\R^n\to \R$, and we write
$p_i(t) = p_i(\mathbf{x}(t))$ for notational simplicity. We emphasize that although the full state $\vc x(t)$ is not available, we assume the observables $p_i$ are 
measurable quantities. These observations may contain certain coordinates of the system state $\vc x$ or, more generally, may be linear or nonlinear functions of the state.

The partial observations $\mathcal U :=\{\vc p(s): s\leq t\}$ up to time $t$ are used to predict the QoI $q(t+\tau)$ at the future time $t+\tau$. 
To this end, we seek to learn a map $N: \mathcal U \to\R$ that predicts the QoI in the future,
\begin{equation}
	\hat q(t+\tau) = N \left[\vc p(s), s\leq t\right].
\end{equation}
We denote the predicted QoI by $\hat q$. In section \ref{MLarch}, we detail the machine learning methods used to learn the map $N$.

In practice, we realize a solution to equation~\eqref{ode1} as a discrete time series obtained by numerical integration or experimental measurements.
For some small time step $\Delta t$, the discrete time series for the state $\vc x(t)$ is  denoted by $\{\vc x^{(0)}, \vc x^{(1)}, ..., \vc x^{(k)},... \}$, where $\vc x^{(k)} = \vc x(k\Delta t)$. Similarly, for the observations $\vc p$ and the quantity of interest $q$, we have the discrete time series 
$\{\vc p^{(0)}, \vc p^{(1)}, ..., \vc p^{(k)},... \}$ and $ \{ q^{(0)}, q^{(1)}, ..., q^{(k)},...\}, $ where $\vc p^{(k)} = \vc p(\vc x^{(k)})$ and $q^{(k)} = q(\vc x^{(k)})$.
\begin{figure*}
	\centering
	\includegraphics[width = 0.9\textwidth]{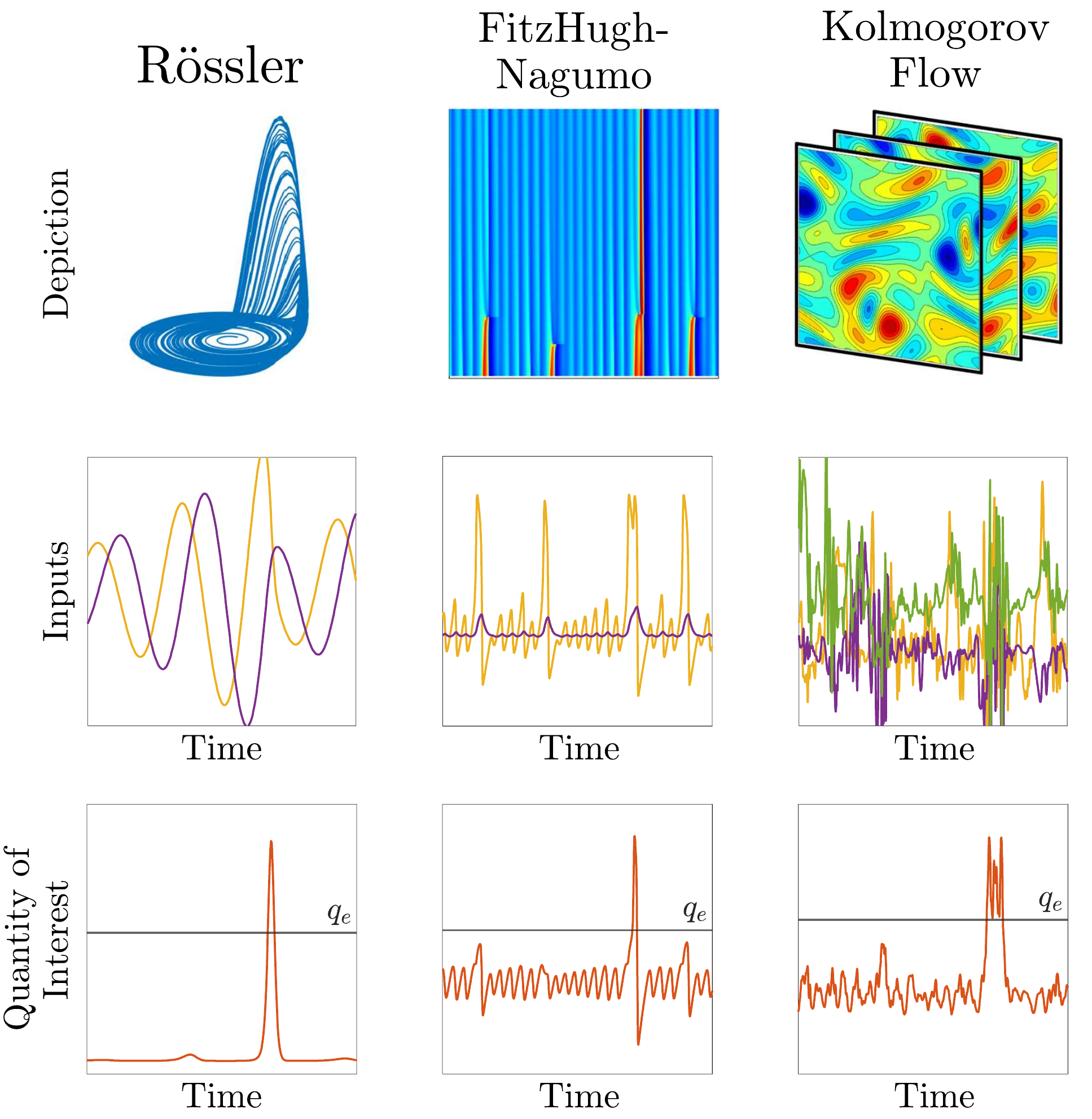}
	\caption{
		The three dynamical systems studied. The complexity of the systems increases from left to right.
	}
	\label{section2Summary}
\end{figure*}

\subsection{{\cb Assumptions and limitations}}
The proposed framework uses simulation data to pre-train a neural network which is subsequently used for prediction from observational data (see figure~\ref{fig:schem}).
Some consistency assumptions have to be made in order to ensure that the pre-trained network is transferable to observational data. For instance, if the distribution of the simulation data is different from the observations, the trained network may fail to accurately predict the out-of-sample events in the observations. To ensure consistency between the simulation data and the observational data, we require the following conditions:
(i) Statistical stationarity: We assume that the system's attractor is statistically stationary so that the distribution of the simulation data and the observations
are in agreement. 
(ii) Constant sampling rate: We assume that the time series $\{\vc p(s) : s\leq t\}$ are sampled at regular intervals $\Delta t$ and that the observational data can be measured with the same frequency. 

Assumption (i) ensures that the dynamics has reached a stationary distribution which is common between the simulation data and the observations. Note that here we assume that dynamical system~\eqref{ode1} accurately models the system; therefore, the only discrepancy between the simulation and observation distributions can arise form observational noise which does not significantly alter the distribution. As a result of assumption (i), the proposed framework is not applicable to non-autonomous systems where the vector field $\vc F$ depends explicitly on time. We note that the system does not necessarily need to be ergodic; we only require that the components of the system attractor are adequately sampled.

Assumption (ii) ensures that the learned map $N$, which maps the input time series $\mathcal U=\{\vc p(s): s\leq t\}$ to the QoI $q(t+\tau)$, is applicable to observations. The map $N$ depends implicitly on the sampling time step $\Delta t$. In other words, a network $N$ that maps input time series sampled at $\Delta t$ intervals is different from the network $N'$ that maps samples which are $\Delta t'$ ($\neq \Delta t$) apart. Therefore, if the simulation data is sampled at a different rate than the observational data, the pre-trained network cannot be applied to make predictions from observations. As a result of assumption (ii), when setting up the training data from simulations, one should ensure that the sampling rate $\Delta t$ matches the frequency of available observations.
We emphasize that the prediction time $\tau$ does not need to be equal to the sampling time step $\Delta t$.

Finally, we note that statistical stationarity is an inherent property of the dynamical system 
whereas constant sampling rate is related to the manner in which the system is observed.

\subsection{Three systems studied} \label{exampleSystems}

We consider three dynamical systems of increasing complexity: the R\"ossler system, the FitzHugh-Nagumo (FHN) system, and the Kolmogorov flow (KF). 
Figure~\ref{section2Summary} shows these systems, their observables and the quantities of interest. Here, we briefly review each system and its significance; a detailed description of each system is provided in Appendix~\ref{appendix_equations}.

\begin{enumerate} 
	\item R\"ossler: The R\"ossler system is a three-dimensional ODE routinely used as a prototypical model for extreme events. 
	Denoting the state variable with $\vc x = (x_1,x_2,x_3)$, the extreme events in this system appear as intermittent chaotic bursts in its $x_3$ component (see figure~\ref{section2Summary}). Here, we consider this system with $x_1$ and $x_2$ components of the state as our observables, and $q=x_3$ as the quantity of interest.
	
	\item FitzHugh-Nagumo: The FitzHugh-Nagumo equations model excitable systems such as neural and cardiac activity.
	Here, we consider a discrete version of the FHN model consisting of $n=101$ diffusively coupled units~\cite{FHNparameters}. 
	Each unit $i$ consists of two variables $(v_i,w_i)$ leading to a 202 dimensional system. We assume only the variables of the first unit $(v_1,w_1)$ are observable. The QoI is the mean voltage $q=(\sum_iv_i)/n$.
	
	\item Kolmogorov flow: The Kolmogorov flow refers to the Navier-Stokes equations with periodic boundary conditions and a sinusoidal shear forcing.
	At high enough Reynolds numbers, the system exhibits extreme events in form of chaotic bursts of the energy dissipation rate~\cite{faraz_adjoint}.
	Here, we consider two sets of observables; one is a particular Fourier mode which was recently discovered to play a major role 
	in the formation of extreme events~\cite{Farazmand2017}. The second set of observables is the vorticity field measured at a few discrete points. 
	We investigate the performance of deep learning methods for each set of observables (i.e., network inputs). In both cases, the QoI is the energy dissipation rate.
\end{enumerate}

% We refer to Appendix~\ref{appendix_equations} for a detailed description of these three systems.

\section{Machine learning architectures} \label{MLarch}
\begin{table*}
	\centering
	\caption{The list of fine-tuned hyperparameters for each neural network.}
	\begin{tabular}{|*{4}{c|}}
		\cline{2-4}
		\multicolumn{1}{c|}{} & \textbf{Feedforward} & \textbf{LSTM} & \textbf{Reservoir Computing} \\ \hline
		\multirow{6}{*}{\rotatebox{90}{\textbf{Hyperparam.}}} &	number of layers     & hidden units &  number of nodes\\ 
		& number of nodes      & number of layers & spectral radius \\ 
		& activation functions &  &  leaking rate\\ 
		& number of delays &  &  input density\\ 
		& delay time&  & reservoir density\\ 
		& &  & regularization weight\\ \hline
	\end{tabular}
	\label{tab:params}
\end{table*}

We use artificial neural networks to predict upcoming extreme events given the available information about the system.  More specifically, we train each neural network to learn the predictor $N$ which
takes partial observations $\mathcal U=\{\vc p(s), s\leq t\}$ from the system as input and returns, as output, the predicted value of the quantity of interest, $\tau$ time units into the future, i.e., $\hat q(t + \tau)$. 

We compare the performance of three deep neural network structures: feedforward (FF), long short-term memory (LSTM), and reservoir computing (RC) networks. 
Each neural network depends on many hyperparameters which have to be prescribed before training the network.
It is well-known that the choice of the hyperparameters can drastically alter the network performance. 
Here, we choose these hyperparameters after an extensive trial-and-error search in the hyperparameter space as listed in Table~\ref{tab:params}.
The optimal set of hyperparameters depends on the system (R\"ossler, FitzHugh-Nagumo, or Kolmogorov flow) as detailed in Appendix~\ref{netarchapp}.
Here, `optimal' refers to the best combination of hyperparameters among those we tested and not optimal over all possible hyperparameters.
In the following, we briefly review each neural network architecture and its hyperparameters.

\subsection{Feedforward neural networks} % XXX 
\begin{figure*}
	\centering
	\includegraphics[width = 0.7\textwidth]{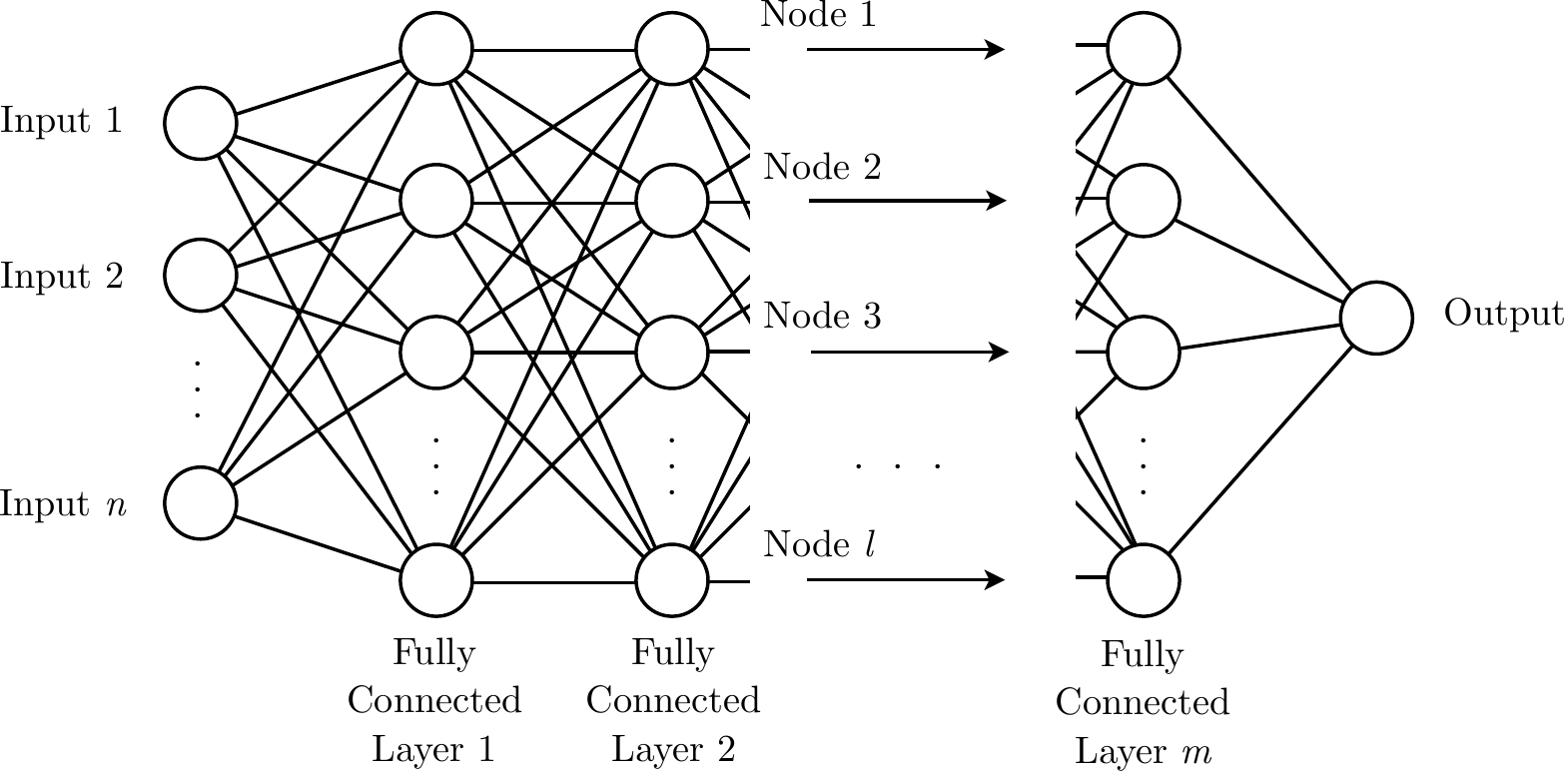}
	\caption{The  FF network architecture with a single output and input data of size $n$.
		There are $m$ layers and $l$ nodes per layer.}
	\label{ffarch}
\end{figure*}

We use a fully connected feedforward neural network as shown in figure~\ref{ffarch}.
Each node in the neural network takes the outputs from the previous layer and transforms them with a nonlinear mapping.
Within a particular layer, the $j$-th node takes the outputs of the previous layer ($x_i$, for $1 \leq i \leq l$), and returns $\sigma(\sum_{i = 1}^l w_{i,j}x_i + b_j)$, where $\sigma: \R \to \R$ is a sigmoidal activation function. 
The weights $w_{i,j}$ and bias $b_j$ are trained using backpropogation and stochastic gradient descent.
This nonlinear composition continues until the network reaches the final (or output) layer. 
The output layer has its own bias $b_o$ and activation function $\sigma_o$. 

%Hornik~\cite{hornik} showed that  FF neural networks are universal approximators for any continuous function. 
%A large enough  FF neural network can produce nearly 100 percent accuracy on the training data. 
%However, in practice, these types of networks often lead to overfitting.
%An overfitted network is unable to generalize well and performs poorly on unseen data. 
%Using a large number of nodes is not always optimal for performance nor computer runtime - training time scales with network size.

We use the \texttt{feedforwardnet} function from MATLAB's Deep Learning Toolbox to implement our FF neural networks for time-series prediction.
We train our networks for $1,000$ epochs using the Levenberg-Marquardt algorithm, a variation of gradient descent, to minimize the loss~\cite{levenberg}. 
The loss function is the mean squared error,
\begin{align}
	MSE = \frac{1}{K}\sum_{i=1}^K(q(t_k+\tau) - \hat{q}(t_k+\tau))^2. \label{mse}
\end{align}
where $q$ (respectively, $\hat q$) denotes the true value of the quantity of interest (respectively, the predicted value of the quantity of interest).
Here, $K$ is the number of predictions made. 
We choose $15\%$ of the original training data as validation data and 
terminate training after validation error increases over $6$ consecutive epochs.

To determine the optimal hyperparameters, we consider between $2$ to $6$ layers with $4$ to $64$ nodes per layer. 
In addition, we consider $\log$-sigmoidal, $\tanh$-sigmoidal, and linear activation functions. 
Two additional hyperparameters correspond to time delays in our input data, as discussed in the following section~\ref{tdintro} below. 
The optimal combination of these hyperparameters for each dynamical system is discussed in Appendix~\ref{netarchapp}.

\subsubsection{Time delay embedding}
\label{tdintro} % XXX 
For FF networks, predicting the QoI $\hat{q}(t+\tau)$ from the instantaneous observations $\vc p(t)$ does not return accurate results. 
Motivated by Takens' embedding theorem~\cite{takens}, we introduce time delays to take into account the history of the observations.   
The time delay embedding collects several previous observations from the time series $\{\vc p(s), \, s\leq t\}$. 
More precisely, for a prescribed number of delays $m\in \mathbb{N}$ and a delay time $s>0$, the time delay embedding $\bm \rho\in \R^{r\cdot m}$ is defined as
\begin{equation}
	\bm\rho(t;m,s) =
	[\vc p(t), \vc p(t-s), \ldots ,\vc p(t-(m-1)s)].
\end{equation}
The special case $m = 1$ corresponds to instantaneous observations, $\bm\rho(t;m,s) = \vc p(t)$.
The function learned using the FF neural network is therefore a map $N: \R^{r \cdot m} \to \R$ from past %the history of the
observations up to the current time $t$
to the predicted quantity of interest $\hat q(t+\tau)$.%, $\tau$ time units into the future.

The values for $m$ and $s$ are chosen differently for each system as detailed in Appendix~\ref{netarchapp}.
The number of delays $m$ depends on the dimension of the system attractor whereas the delay time $s$ depends on the decorrelation time of the system. 
Arbitrarily increasing $m$ is unfeasible due to training time constraints and $s$ must be chosen carefully so that there is little redundancy between delayed observations.

LSTM and RC are recurrent neural networks which implicitly take the history into account. As a result, time delay embedding is unnecessary for training LSTM and RC networks, and
is only used for training the FF neural network. In fact, we have observed that including the time delays in the input data explicitly tends to deteriorate the performance of the LSTM and RC networks.

%Subramanian et al. 2019~\cite{subramanian2019} found that  FF networks performed worse than LSTM and RC networks for predictions with limited information. 
%This performance disparity might be expected since the latter two network architectures implicitly incorporate past information. 
%We augment  FF networks with time delays to try to close the performance difference. 

\subsection{Long short-term memory neural networks}
\label{sec:lstm}
Long short-term memory (LSTM) networks are a particular type of recurrent neural network (RNN) which use gates
to control the effect of past history on the current state. 
This construction allows them to capture long-term dependencies in the time series~\cite{Schmidhuber1997, Sherstinsky2018}.
\begin{figure*}
	\centering
	\subfigure[]{\includegraphics[width = .7\textwidth]{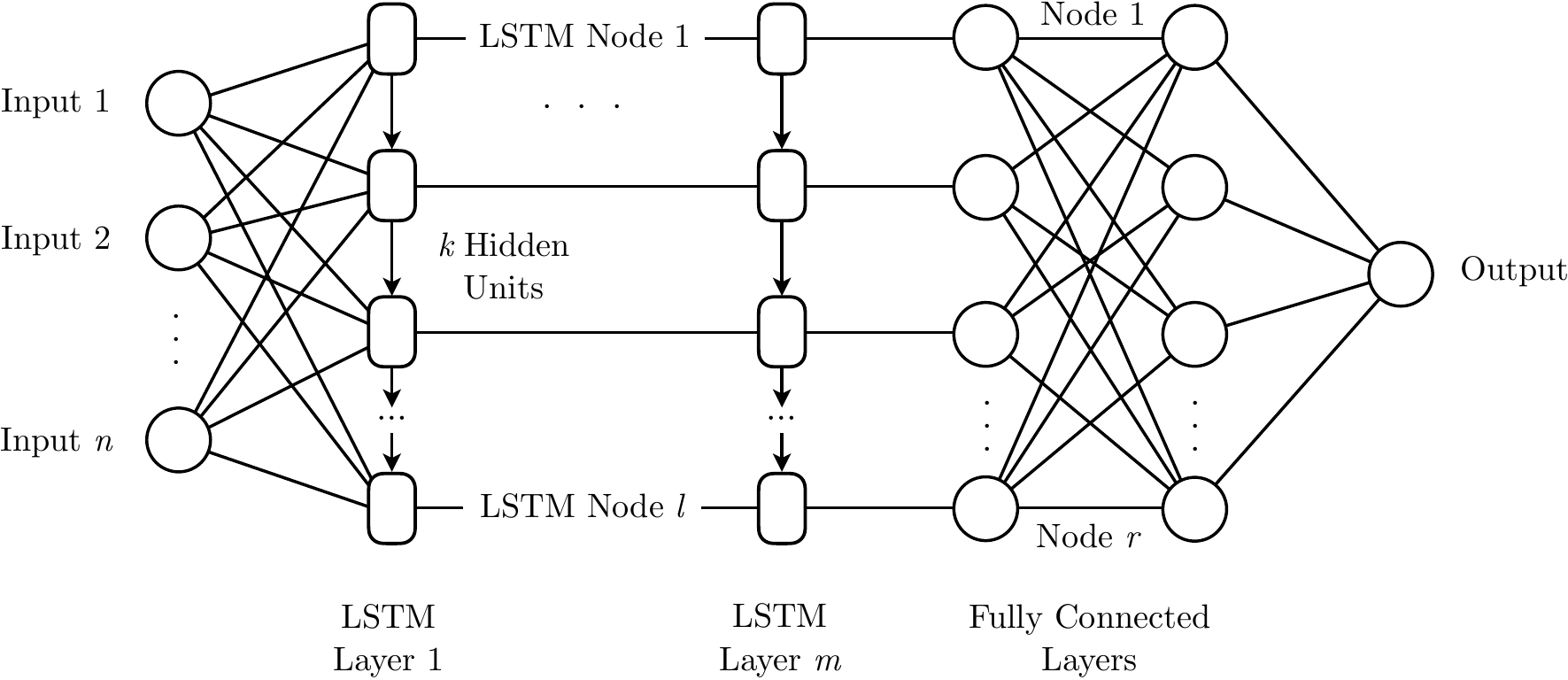}}
	\subfigure[]{\includegraphics[width = 0.2\textwidth]{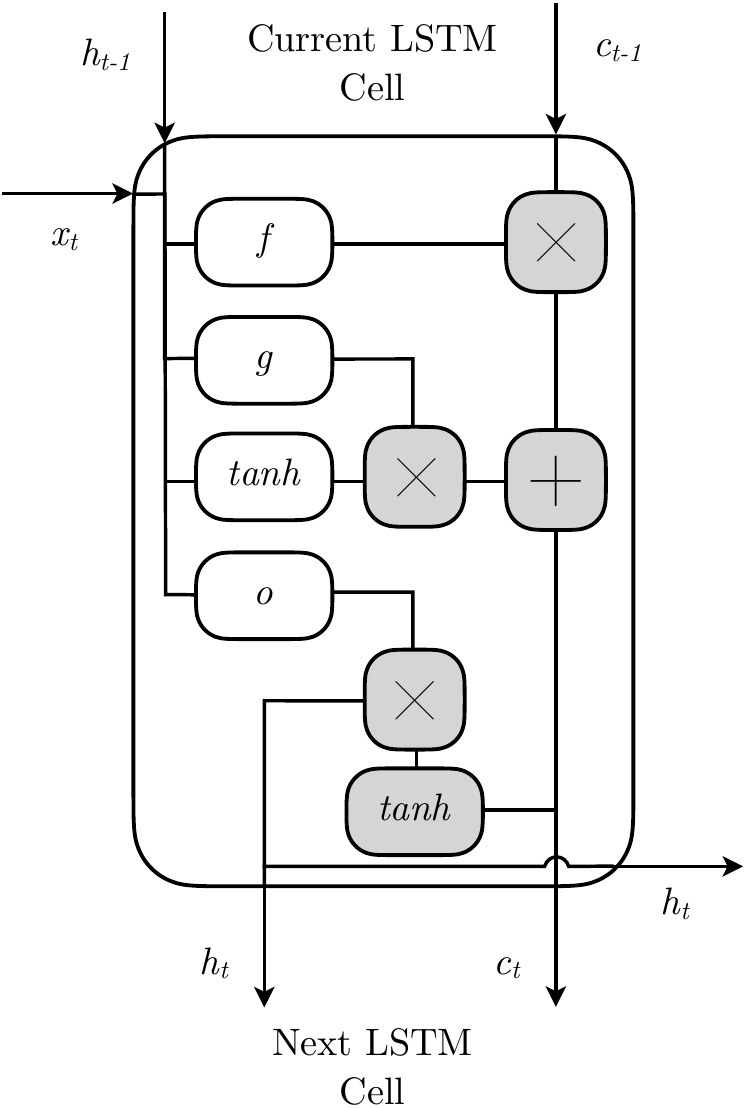}}
	\caption{(a) LSTM architecture with $m$ layers followed by a fully connected network. (b) An LSTM cell.}
	\label{lstmarch}
\end{figure*}

Figure~\ref{lstmarch} illustrates our LSTM architecture consisting of LSTM layers followed by a fully connected network.
The networks use an input layer of dimension $n$, followed by $m$ LSTM layers, followed by a fully connected layer ending in an output layer of dimension one.
The fully connected and output layers in figure~\ref{lstmarch} are identical to their  FF neural network counterparts; the new LSTM layers enable the network to “remember” past history.

Each LSTM layer takes a time series and, after iterating through each time step, returns a reconstructed time series composed of significant details the network remembered through training. % ??? 
At each time step $t$ the cell combines the time step information with the remembered information. 
The previous cell determines the remembered information and the modifications to make, known as the cell state $c_{t - 1}$ and the hidden state $h_{t - 1}$, respectively. 
The number of modifications made, known as hidden units, is a hyperparameter.

Each LSTM node uses three gates to decide which information should be retained or discarded in producing the output (see figure~\ref{lstmarch}(b)). The forget gate $f$ determines which information the LSTM block removes from the layer's memory.
The input gate and cell candidate, denoted by $\tanh$ and $g$, determine what information from the input data $x_t$ the LSTM block incorporates to update the cell state in the LSTM node. The output gate $o$ determines how the cell state $c_{t-1}$ contributes to the output relative to $h_{t-1}$ and $x_t$. 

% In the training process, the weights and biases in the fully connect layer are 
Each gate in the LSTM node has two weight matrices and a bias.
One weight matrix is associated with the input $x_i$ while the other weight matrix is associated with the hidden state $h_{t-1}$.
In the training process, the weights and biases of both the LSTM network and the fully connected network are trained by backpropagation. 
% We note that during training the ordering of data is important -- it must be order in the same way as the time series was generated. 
% To train the LSTM network appropriately, the time series of observations is 

We implement LSTM networks for sequence-to-sequence regression using MATLAB's Deep Learning Toolbox.
% Network architecture ranges from $1$ to $3$ LSTM layers and $8$ to $200$ hidden units per layer. 
The two hyperparameters for LSTM networks we adjust are the number of layers and the number of hidden units per layer.
% We choose between $1$ and $3$ layers and $8$ to $200$ hidden units per layer. 
We explore hyperparameter values between $1$ and $3$ layers and $8$ to $200$ hidden units per layer. Each network trains for $250$ epochs using the Adam algorithm to minimize the loss function~\eqref{mse}.

\subsection{Reservoir computing}
Reservoir computing networks are the second RNN we consider here.
The key structure is a reservoir of nodes with prescribed connections capturing complex temporal dynamics~\cite{schrauwen2007}.  
Our RC network takes an input of dimension $n$ and outputs a scalar. 

\begin{figure*}
	\centering
	\includegraphics[width = 0.8\textwidth]{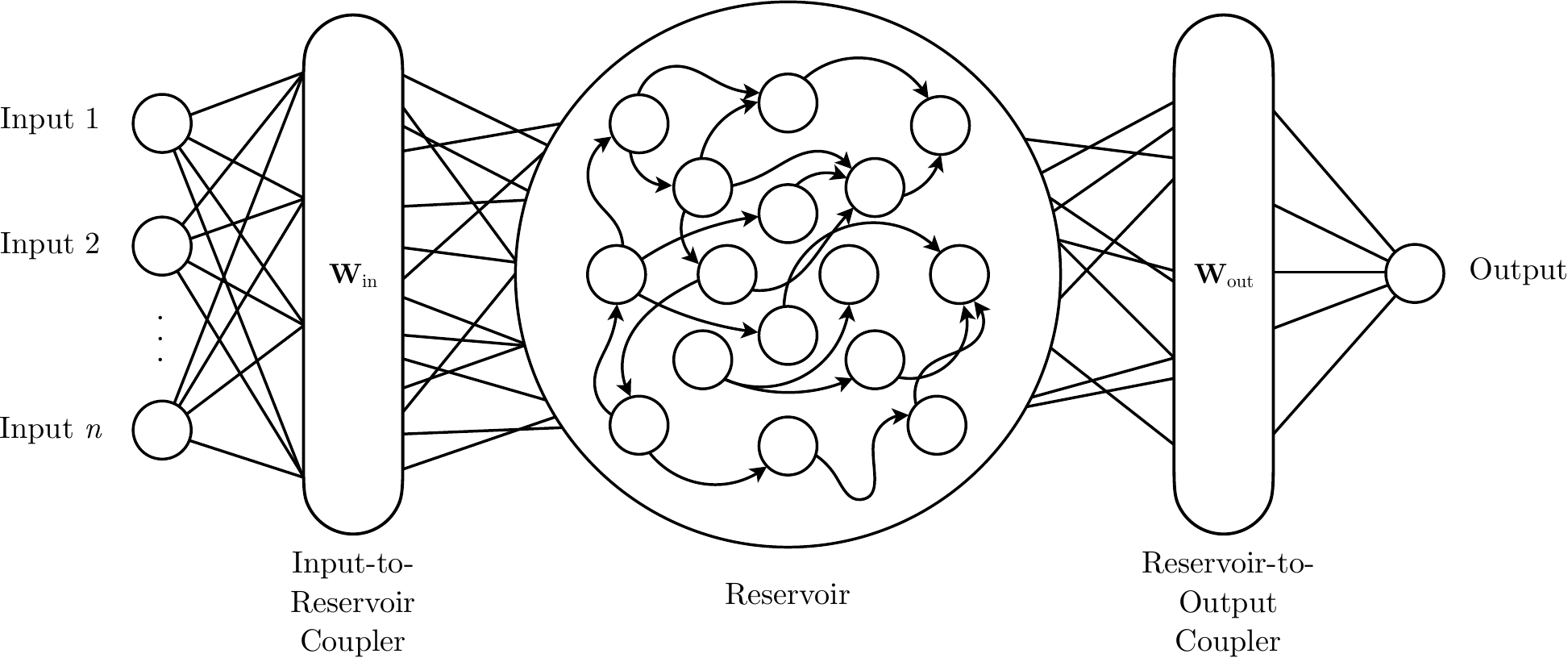}
	\caption{Reservoir computing network with $n$ inputs and 1 output. }
	\label{fig:rcarch}
\end{figure*}

As shown in figure~\ref{fig:rcarch}, the reservoir architecture consists of two couplers and a high dimensional dynamical system. 
The input-to-reservoir (I/R) coupler maps the input data to the reservoir while the reservoir-to-output (R/O) coupler maps the reservoir state to an output. 
% A reservoir computer consists of two mappings and a
% : an input-to-reservoir ($I/R$) coupler and a reservoir-to-output ($R/O$) coupler, and a high dimensional dynamical system $R$ called the reservoir .
% The $I/R$ coupler maps an input $\vc p \in \R^p$ into the reservoir with $\win:\R^m\to \R$. 
Here both couplers are chosen to be linear mappings.
The I/R coupler uses $\win\in\R^{m\times n}$, whose entries are sampled from a uniform distribution $U(-a, a)$ for some $a>0$, to map the observations $\vc p(t)$ into the reservoir,
\begin{equation}
	\vc u(t)=\win \vc p(t),
\end{equation}
where $\vc u(t)$ is used to update the reservoir state $\vc r(t)$.

The entry $r_j(t)$ of the reservoir state $\mathbf{r}(t)\in\mathbb R^m$ corresponds to the $j$-th reservoir node.
The reservoir is a directed Erd\"os–Rényi network with $m \times m$ adjacency matrix $\mathbf{A}$ describing connections between nodes. 
We denote the spectral radius of $\mathbf{A}$ by $\rho \in \R$. 
The reservoir density is the percentage of nonzero entries in $\mathbf{A}$. The initial reservoir state $\mathbf{r}(0)$ is chosen at random. 
Within the reservoir, the reservoir state is updated by
$\mathbf{r}(t + \Delta t) = \tanh (\mathbf{Ar}(t) + \mathbf{u}(t))$. 
The reservoir then passes the updated reservoir state $\vc r(t+\Delta t)$ through the R/O coupler via $\hat q(t+\tau)=\wout \mathbf{r}(t) \in \R$, where $\wout\in \R^{1\times m}$ denotes the output weights. 
The vector $\wout$ is optimized during training to minimize the least squares error, with an $L_2$ regularization term.  

We create RC networks using \texttt{easyesn} library in Python~\cite{easyesn}. 
As in the case of the LSTM network, our inputs do not include time delay embedding since the recurrent nature of the RC network implicitly accounts for the history.

RC has several hyperparameters. We vary the input density, or the portion of nonzero entries in $\mathbf{W_{\text{in}}}$ between $0.1$ and $1$.
Within the reservoir, we vary the number of nodes from $100$ to $5,000$ and the spectral radius $\rho$ of the adjacency matrix $\vc A$
from $0.1$ to $1$. 
The penalization weight associated with the $L_2$ regularization is a hyperparameter, varied from $10^{-5}$ to $1$.
Finally, a reservoir computer's leaking rate $\ell$ determines how frequently the reservoir updates during training: $\ell^{-1}$ times per time instance $t$.
We also vary the leaking rate from $0.1$ to $1$. The optimal value of the hyperparameters for each dynamical system is discussed in Appendix~\ref{netarchapp}.

\section{Results and discussion} \label{results}
In this section, we use deep learning to predict extreme events in three dynamical systems: 
R\"ossler (Section~\ref{sec:results_rossler}), 
FitzHugh-Nagumo (Section~\ref{sec:results_FHN}), and 
Kolmogorov flow (Section~\ref{sec:results_kolm}).
For each system, we generate a long-term numerical simulation which returns the full state $\vc x(t)$. 
However, to mimic applications where the full state is not available, we do not use $\vc x(t)$ directly for training the neural networks. 
Instead, we use the full state information $\vc x(t)$ to generate the corresponding time series of the observables $\vc p(t)$, as depicted in figure~\ref{fig:schem}.

\begin{figure*}
	\centering
	\includegraphics[width = .9\textwidth]{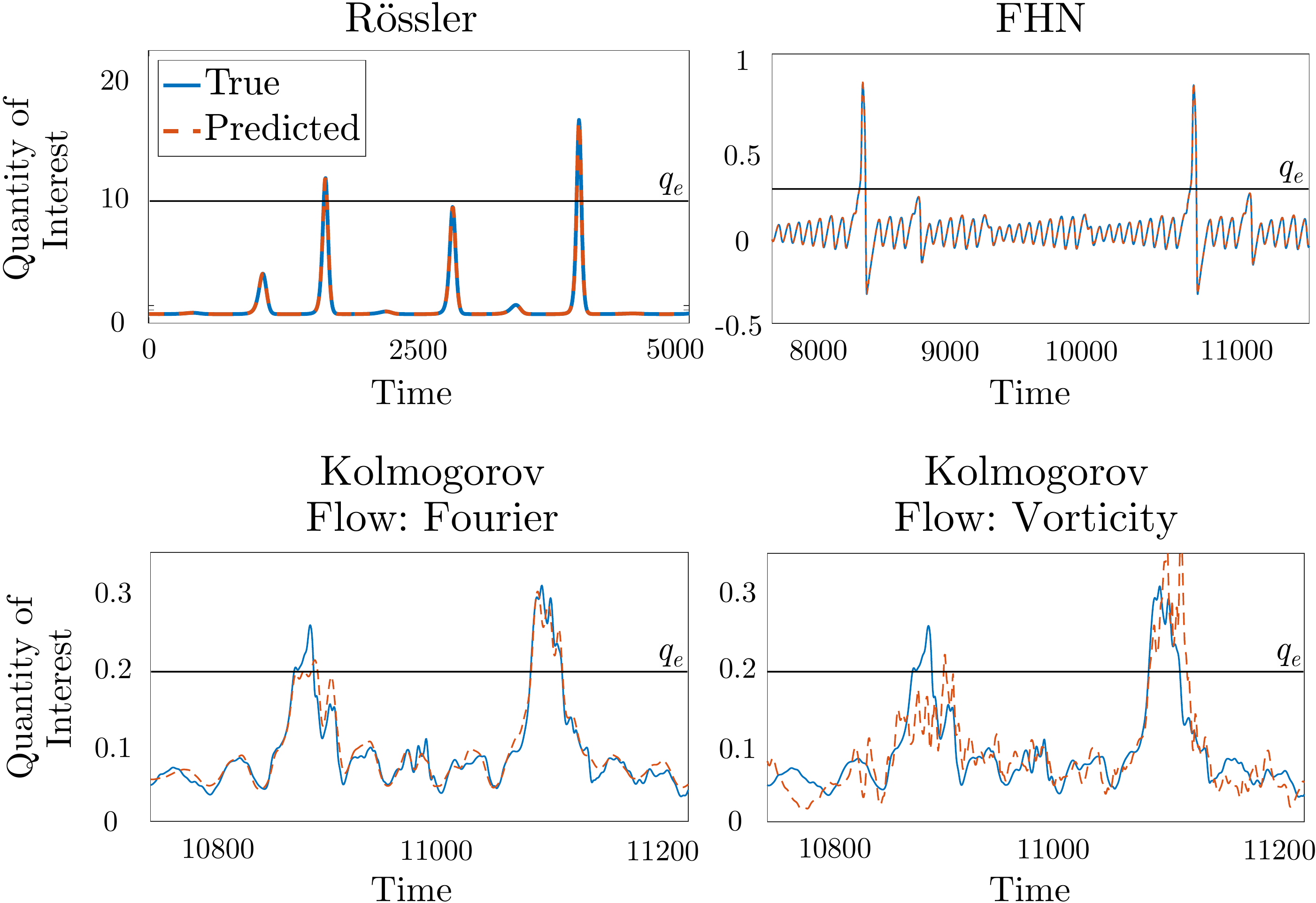}
	\caption{Close-up views of the time series for each system's quantity of interest $q(t)$. 
		The blue curves show the true value $q(t)$, while the dashed lines mark the best prediction $\hat q(t)$.}
	\label{sec4_timeseries}
\end{figure*}

The first $75\%$ of the observable time series is used for training the neural network and the remaining $25\%$ is used to test the trained network.
For each system, we quantify the error between each system's QoI $q$ and the corresponding neural network prediction $\hat q$ using two 
measures of accuracy described in section~\ref{sec:accuracy}.
For each network structure outlined in Section~\ref{MLarch}, we optimize the hyperparameters for each dynamical system separately.
We present the optimal hyperparameter combinations that we use to train each neural network structure in Appendix~\ref{netarchapp}. 

The produced observable time series $\vc p(t)$ only contain round-off and numerical discretization errors which are negligible.
In practice, however, observations are always polluted with noise. To mimic the observational noise, we artificially add noise to our synthetic test 
data. Noise is added proportionally to each component of the observation $\vc p(t)$ at four intensities: $0\%$, $5\%$, $10\%$, and $20\%$. 
More precisely, let $\sigma_i$ denote the standard deviation of the time series of the $i$-th observation $p_i(t)$.
Then the corresponding noisy data $\tilde p_i(\vc x(t))$ is given by
\begin{equation} 
	\tilde p_i(\vc x(t)) = p_i(\vc x(t)) + \alpha \xi_i(t), \quad \xi_i(t) \sim \mathcal{N}(0,\sigma_i ),
	\label{eq:noise}
\end{equation}
where $\xi_i(t)$ is a realization of $\mathcal{N}(0,\sigma_i)$, the normal distribution with mean 0 and standard deviation $\sigma_i$. 
The parameter $\alpha\in \{0,0.05, 0.10, 0.20\}$ controls the noise intensity with $\alpha=0$ corresponding to the clean data obtained from numerical simulations.

Figures \ref{sec4_timeseries} to \ref{sec4_noiseTest} contain our main results. All reported results correspond to the test data.
Figure~\ref{sec4_timeseries} displays samples of the predicted times series versus their true values.
Figure~\ref{sec4_summary} illustrates the predictive power of the deep learning architectures for each dynamical system. Figure~\ref{sec4_tau} shows the effect of varying the prediction time $\tau$ and figure~\ref{sec4_noiseTest} displays how noise affects prediction accuracy.
In sections \ref{sec:results_rossler}-\ref{sec:results_kolm}, we describe these results in detail for each dynamical system. 
But first, in Section~\ref{sec:accuracy}, we describe two quantities used to measure the prediction accuracy of the trained neural networks.

\subsection{Quantifying prediction accuracy}\label{sec:accuracy}
In order to quantify the accuracy of our extreme event predictions, we use two measures: normalized root-mean square error and 
area under the precision-recall curve. The former simply quantifies the accuracy of time series predictions, while the latter 
is more suitable for quantifying the accuracy of extreme event predictions.
\begin{figure*}
	\centering
	\includegraphics[width = .9\textwidth]{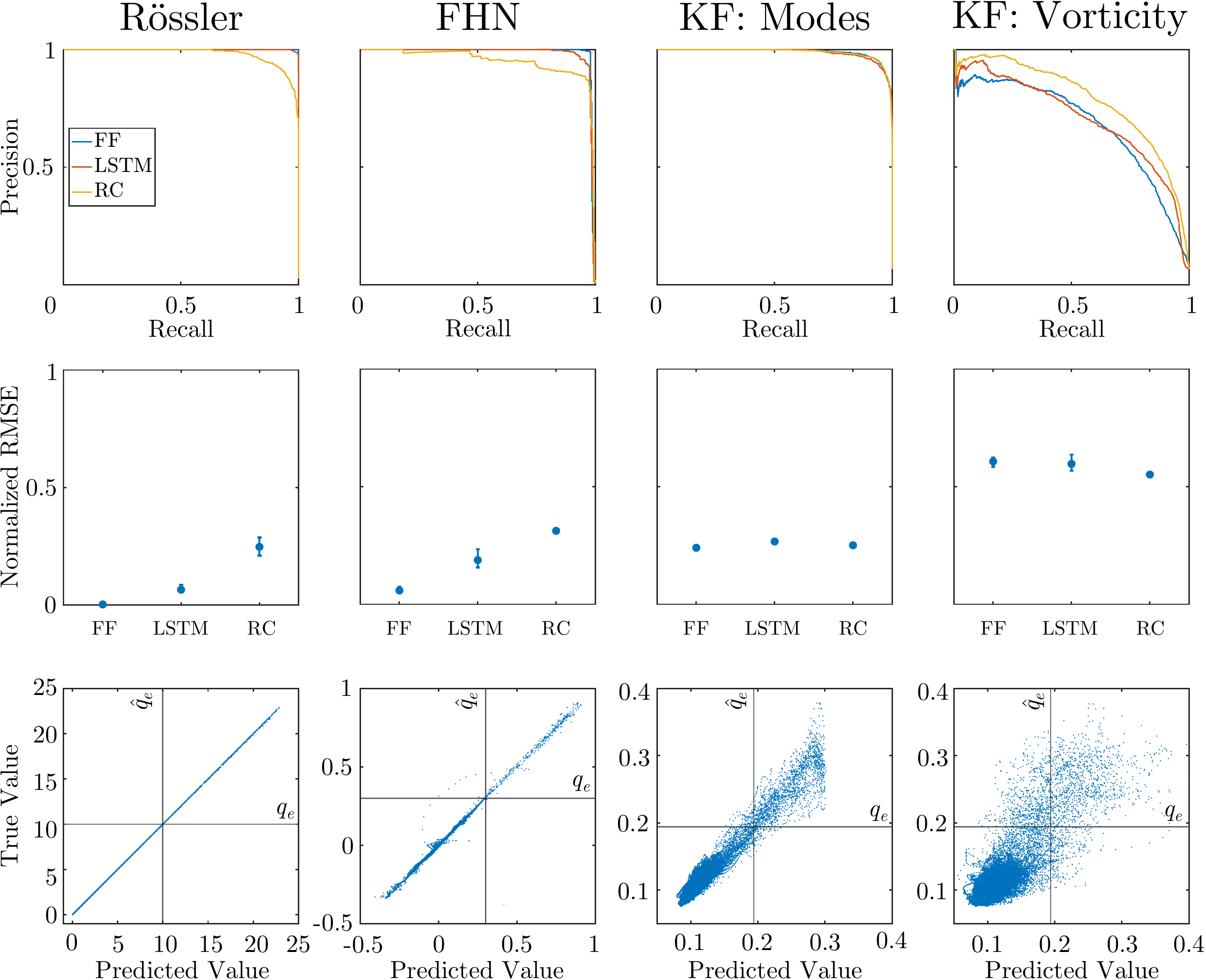}
	\caption{Summary of neural network performance for each system with noiseless data. 
		The precision-recall curves (first row) quantifies the prediction skill of each network.
		The NRMSE plots (second row) quantify the fit of the predicted QoI, with error bars describing the maximum and minimum NRMSE for repeated training of the networks. 
		The true vs. predicted plots (third row) correspond to the best network architecture for each system.}
	\label{sec4_summary}
\end{figure*}

We define the normalized root-mean-squared error (NRMSE), 
\begin{align}\label{NRMSE}
	\text{NRMSE} = \frac{\left[\frac{1}{K}\sum_{k=1}^K (q(t_k+\tau) - \hat q(t_k+\tau))^2\right]^{1/2}}{\sigma_q},
\end{align}
where $\sigma_q$ is the standard deviation of the true QoI, $\{q(t_k+\tau)\}_{k=1}^K$.
Note that NRMSE is the ratio of the square root of loss function~\eqref{mse} and the standard deviation $\sigma_q$.
The normalization by $\sigma_q$ allows us to make base comparisons across different dynamical systems.

NRMSE measures the deviation in our predictions from the true value with no consideration given to extreme events.
As reviewed in Ref.~\onlinecite{sapsis2019},
there are measures of accuracy which are better suited for extreme event prediction.
Here, we quantify the accuracy of extreme event prediction using area under the precision-recall curve, or area under curve (AUC) for short. 
Recall that an extreme event is registered if $q>q_e$ where $q_e$ is the extreme event threshold. 
As a result, the predicted QoI $\hat q$ may not coincide with $q$, but as long as $\hat q>\hat q_e$, the extreme event is correctly predicted.
Here $\hat q_e$ is the extreme event threshold according to the predicted QoI; in practice, $\hat q_e$ is close to the true extreme event threshold $q_e$, 
but the two do not necessarily coincide.

Each prediction $\hat q$ is classified in one of four ways: 
1. True positive (TP): A true positive corresponds to a successful prediction of a true extreme event, i.e., $q>q_e$ and simultaneously $\hat q>\hat q_e$.
2. True negative (TN): A true negative corresponds to a correct prediction that no extreme event will take place, i.e.,  $q<q_e$ and simultaneously $\hat q<\hat q_e$.
3. False positive (FP): A false positive corresponds to the case where an extreme event was predicted but did not actually occur, i.e., $q<q_e$ but $\hat q>\hat q_e$.
4. False negative (FN): A false negative corresponds to the case where an extreme event occurred but the neural network failed to predict it, i.e., $q>q_e$ but $\hat q<\hat q_e$.

Precision and recall quantify extreme event prediction accuracy with regards to these four classifications.
These quantities are defined by
\begin{subequations}
	\begin{equation}\label{eqn:precision}
		\text{precision}(\hat q_e) = \frac{\text{TP}(\hat q_e)}{\text{TP}(\hat q_e) + \text{FP}(\hat q_e)},
	\end{equation}
	\begin{equation}\label{eqn:recall}
		\text{recall}(\hat q_e) = \frac{\text{TP}(\hat q_e)}{\text{TP}(\hat q_e) + \text{FN}(\hat q_e)}.
	\end{equation}
\end{subequations}

For a given prediction threshold $\hat q_e$, precision measures the ratio of successfully predicted extreme events to the total number of predicted extreme events whereas recall measures the ratio of successfully predicted extreme events to the total number of true extreme events. 
Since extreme events are rare, precision and recall more accurately capture the prediction skill than TP and TN rates alone~\cite{davis2006, sapsis2019, saito2015}.
In the best case scenario, where no false predictions are made, precision and recall are both 1.

We combine both these quantities to create the precision-recall curves, parameterized by the threshold $\hat{q}_e$. 
Varying the threshold $\hat q_e$ allows us to examine how precision and recall change under a moving goalpost. 
Note that, if $\hat q_e$ is very small, most extreme events will be correctly predicted leading to a large number of TPs, but at the same time there will be a large number of FPs resulting in low precision. Conversely, if $\hat q_e$ is too large, then the number of FN predictions will be large resulting in low recall. A reliable predictor must have
precision and recall close to 1 for a wide range of prediction thresholds $\hat q_e$. One way to quantify this range is to compute the area under the precision-recall curve, or AUC for short. 
The AUC is the area the precision-recall curve encompasses in the unit square ranging between 0 and 1, with more skillful predictors having an AUC closer to 1. 
The first row of figure~\ref{sec4_summary} provides examples of precision-recall curves.  

\subsection{R\"ossler system}\label{sec:results_rossler}
The R\"ossler system is the least challenging system discussed here due to its lower dimensionality and relatively simpler dynamics. 
This system consists of three state variables $\vc x =(x_1,x_2,x_3)$ where the coordinate  $x_3$  exhibits chaotic bursts (see figure~\ref{sec4_timeseries}). 
Therefore, we take $q=x_3$ as the quantity of interest with the extreme event threshold $q_e=10$. The other two coordinates are used as observables, $p_1=x_1$ and $p_2=x_2$. We refer to Appendix~\ref{app:rossler} for a detailed description of the R\"ossler system. In Appendix~\ref{app:rossler_param}, we discuss optimal hyperparameters used for each network when trained with R\"ossler data.

{\cb To generate training and testing data, we integrate the system for $500$ time units with the results recorded every $\Delta t=0.05$. Increasing the sampling time to $\Delta t=0.5$ does not significantly alter the reported results, indicating robustness to moderate changes to the sampling rate. 
However, increasing the sampling rate further will eventually reduce the prediction skill of the networks across all three dynamical systems. In R\"ossler, for instance, increasing the sampling time beyond $\Delta t=2$ led to a significant drop in the corresponding AUC.}
\begin{figure}
	\centering
	\includegraphics[width=.4\textwidth]{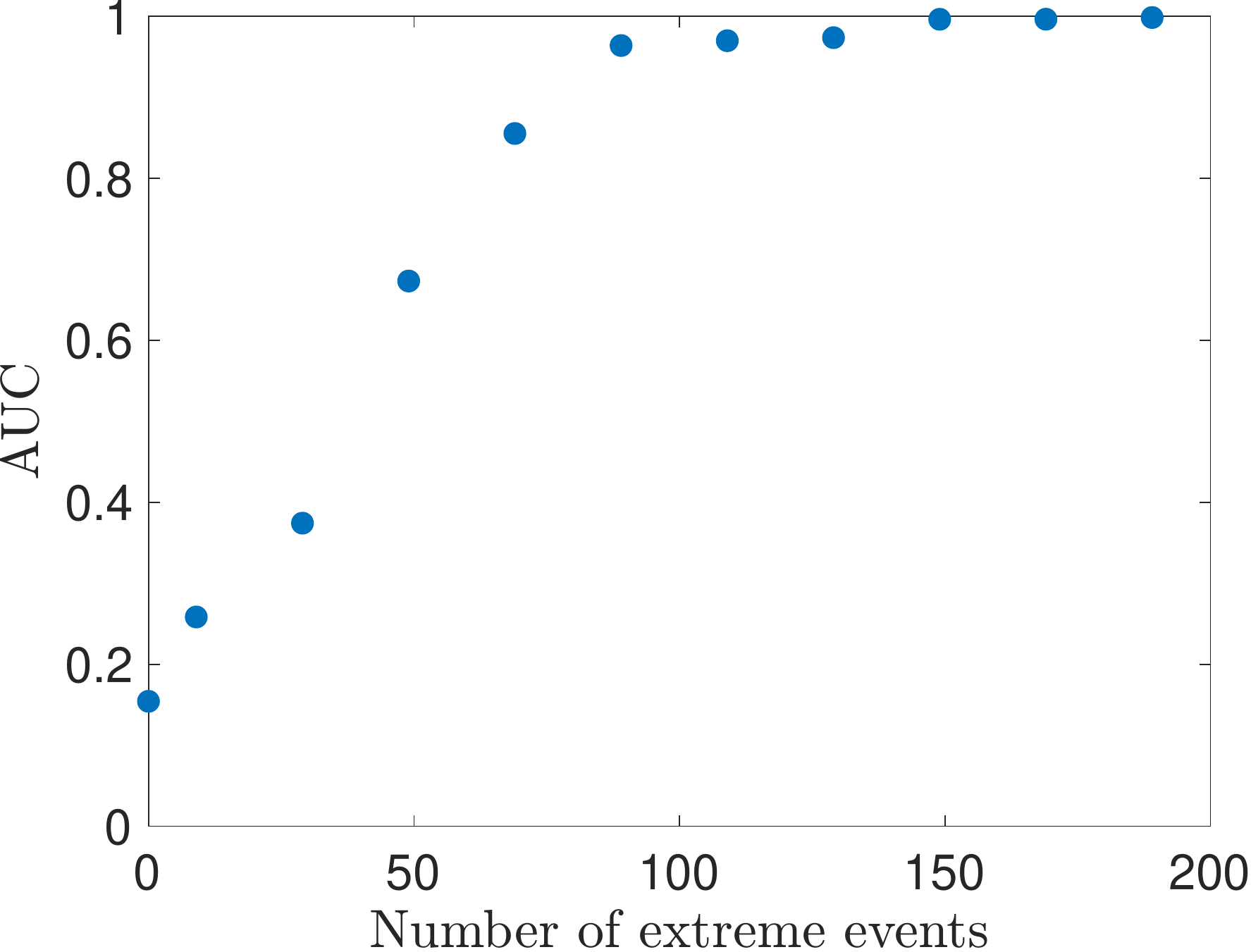}
	\caption{{\cb AUC for the R\"ossler system as the number of extreme events in the training data increases.}}
	\label{fig:NumbEE}
\end{figure}

{\cb The amount of required training data is not a priori known. However, figure~\ref{fig:NumbEE} gives an indication as to when enough training data from simulations is obtained. As the number of extreme events in the training data increases, the AUC initially increases rapidly. After the training data is long enough to contain about 100 extreme events, the AUC plateaus near 1, indicating that enough training data has been gathered and therefore we stop the numerical simulations. We also observed a similar plateauing behavior for the FitzHugh--Nagumo system and the Komogorov flow, although the required number of extreme events before reaching the plateau depends on the system.}

\begin{figure*}
	\centering
	\includegraphics[width = .9\textwidth]{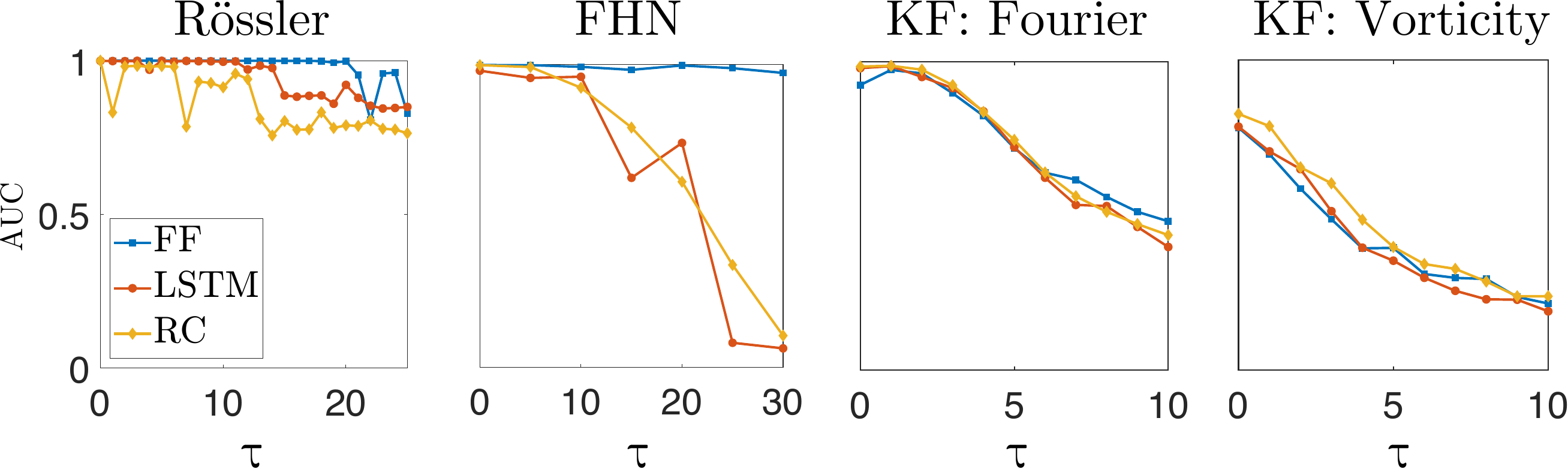}
	\caption{
		AUC as a function of the prediction time $\tau$.}
	\label{sec4_tau}
\end{figure*}

\begin{figure*}
	\centering
	\includegraphics[width = .9\textwidth]{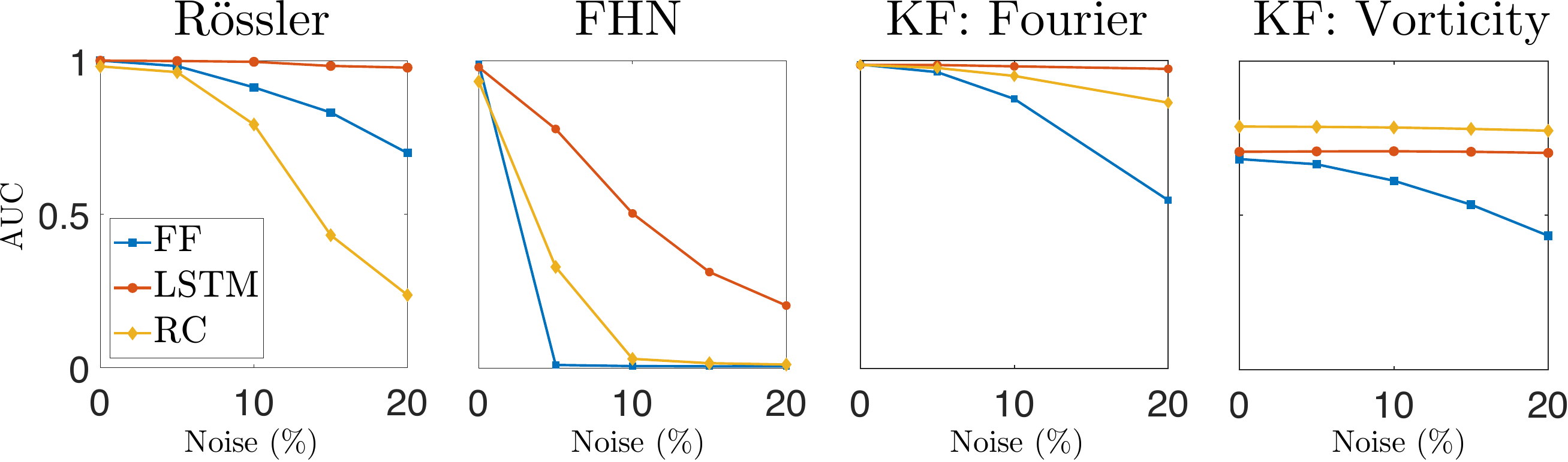}
	\caption{Effect of noise on predictions.
		Each panel shows how the area under the precision-recall curve (AUC) changes as the noise intensity increases. 
		Noise is added only to the testing data. 
	}
	\label{sec4_noiseTest}
\end{figure*}

We first discuss the neural network predictions in the absence of observational noise.
As shown in the first column of figure~\ref{sec4_summary}, 
all three networks generate nearly exact predictions on testing data in the absence of noise, each with an AUC above $0.98$.
The RC network performs slightly worse at an AUC of $0.98$ whereas the AUC for Feedforward and LSTM networks is nearly equal to $1.00$.

Training the network weights is not generally a convex optimization problem. Therefore, retraining the network may lead to different results. In order to quantify the reproducibility of the results, we retrain each network 10 times with different initial network weights.
The error bars in the NRMSE plot within figure~\ref{sec4_summary} correspond to the minimum and maximum value over ten repeated training of each network. The error bars are relatively small across all systems and all networks, demonstrating robustness under retraining.
For R\"ossler, the feedforward network has the best average NRMSE at $0.003 \pm 0.0002.$ 
LSTM networks perform slightly worse with an NMRSE within $0.098 \pm 0.0109$. Finally, the RC network has the largest NRMSE within the range $0.248 \pm 0.040$.

{\cb We also quantify the prediction skill as a function of the prediction time $\tau$. Because of sensitivity to initial conditions, the predictability horizon of a chaotic system is always limited and often scales inversely with the leading Lyapunov exponent of the system. This limitation also applies to extreme event predictions~\cite{PRE2016}.  However, as we explained in Section~\ref{sec:accuracy}, prediction of extreme events is a more forgiving task as compared to time series prediction.}
For the R\"ossler system, increasing the prediction time $\tau$ shows an expected decrease in accuracy for all networks, as shown in figure~\ref{sec4_tau}.
Feedforward networks maintain the AUC at nearly $1$ until $\tau = 20$, while the LSTM network maintains the AUC at nearly $1$ until $\tau = 12$.
The RC networks perform worse than FF and LSTM networks as $\tau$ increases.

{\cb Next we examine the effect of observational noise on the predictions. Note that, although observations are always polluted with noise, 
the training data in our framework is gathered offline using numerical simulations with no significant noise (see figure~\ref{fig:schem}). 
Nonetheless, noise can be artificially added to the simulation data to mimic the observational noise. One may argue that the networks should be trained 
on clean simulation data (with no artificial noise) and hope that the network will perform well under small to moderate amounts of observational noise.
In the following, we show that this approach is ill-advised since adding some artificial noise in the training phase often improves the network performance.

To this end, we first consider the case where no noise is added to the training data and only the testing data is polluted with noise as described in equation~\eqref{eq:noise}.}
As shown in figure~\ref{sec4_noiseTest}, we see notable differences in accuracy among the network architectures. 
LSTM networks consistently maintain their predictive power with the AUC decreasing from nearly $1.00$ to $0.98$ as the noise intensity increases from $0\%$ to $20\%$. 
Feedforward networks perform similarly for $0\%$ and $5\%$ noise; but as the noise intensity increases to $20\%$, the AUC decreases to $0.7$. 
Noise affects the RC network the most, with the AUC dropping to $0.79$ for $10\%$ noise. As the noise intensity reaches $20\%$, the AUC of RC decreases significantly to $0.24$.

{\cb The detrimental effect of noise is more pronounced for the FitzHugh--Nagumo system. We show in Section~\ref{sec:results_FHN} that adding artificial noise to the training data, as well as the testing data, significantly improves the prediction results (see figure~\ref{sec4_FHN_noiseTrain}).}

\subsection{FitzHugh--Nagumo system}\label{sec:results_FHN}
The FitzHugh--Nagumo system is a network of chaotic oscillators composed of coupled units. 
The coordinates of the $i$-th unit are denoted with $(v_i,w_i)$.
Here, we consider a model with 101 completely coupled units resulting in a 202 dimensional system of ODEs.
We assume that only the coordinates of the first unit are observable, so that $p_1=v_1$ and $p_2=w_1$.
The quantity of interest is the average of the $v_i$ coordinates,
$q = \left(\sum_{i=1}^n v_i\right)/n$.
The FHN oscillators occasionally synchronize so that the $v_i$ coordinates align, leading to intermittent bursts in the average $q$ (see figure~\ref{sec4_timeseries}). The extreme event threshold is $q_e=0.3$.

We integrate the FHN system for $2\times 10^5$ time units with the results saved every $\Delta t=1$ time unit.
The FHN model is described in detail in Appendix~\ref{app:FHN}. The optimal hyperparameters for training the neural networks are detailed in Appendix~\ref{app:FHN_param}.

Examining the second column of figure~\ref{sec4_summary}, the precision-recall curves lie in the top right yielding large AUC values. 
The  FF network performs best with an AUC of $0.99$ while the LSTM network had a comparable AUC of $0.98$. The reservoir network has the smallest AUC of $0.93$. 
The scatter plot lies close to the diagonal line indicating the predicted quantity of interest closely matches its true value, although the predictions are not as accurate as the R\"ossler system. This is to be expected since FHN is higher dimensional and only two out of its 202 coordinates are observed.

With multiple network trainings, the FF network performs the best with NRSME at $0.059\pm 0.014$. 
The LSTM network has the largest variability in NRMSE which lies in $0.188\pm 0.039$.
Although the reservoir network has the largest NRMSE, the results have the most consistency, with NRMSE within $0.312\pm0.09$. 
Recall that, for the R\"ossler system, the FF and LSTM networks were the most robust under retraining. In contrast, for the FHN system, the FF and RC networks are the most robust.

As shown in figure~\ref{sec4_tau}, the prediction accuracy decreases as prediction time increases.
However, there is a significant loss in predictive power for the LSTM and RC networks, with
the AUC declining to near 0 for the LSTM and RC networks at $\tau = 25$. 
In contrast, the FF network maintains accurate predictions, with an AUC of 0.97 for the prediction time $\tau=30$.

For the FHN system, an interesting phenomenon takes place when adding noise to the testing data.
As shown in figure~\ref{sec4_noiseTest}, the results are extremely sensitive to noise.
For instance, for the FF network the AUC is $0.99$ when using noise-free testing data; but even adding $5\%$ noise to the testing data, 
the AUC drops to $0.01$. 
While the LSTM network appears slightly more robust to noise, its predictions on data polluted with $20\%$ noise are still inaccurate, with the AUC dropping to around $0.25$.  Contrast this with the LSTM results on the R\"ossler system where the AUC remains close to 1 even when $20\%$ noise is added to the testing data.

\begin{figure*}
	\centering
	\includegraphics[width=.95\textwidth]{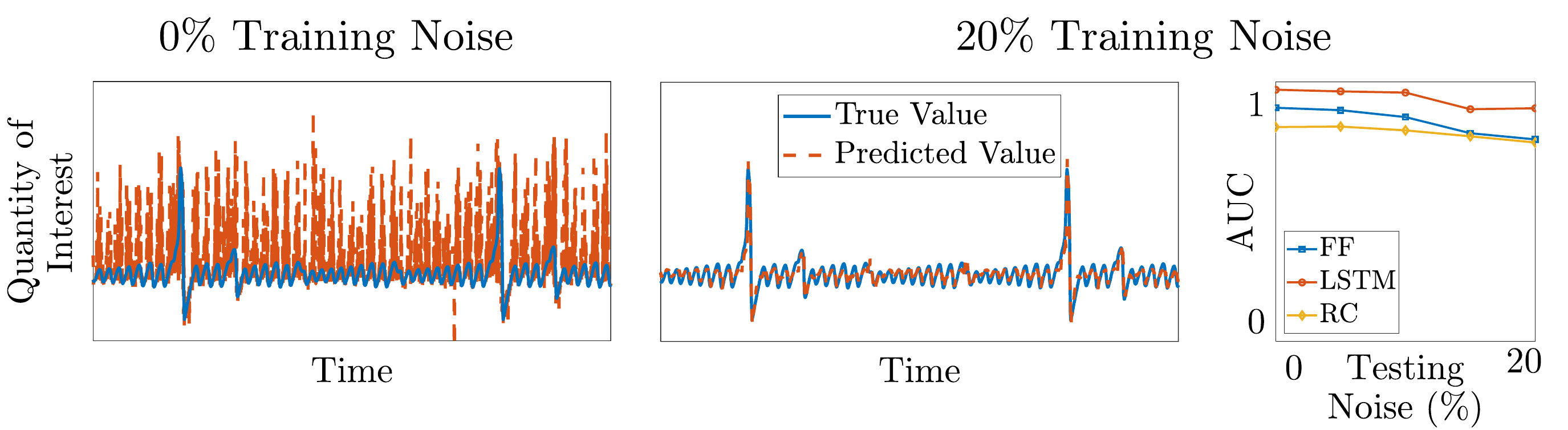}
	\caption{Effect of training noise on predictions for the FitzHugh-Nagumo system. 
		The feedforward network is trained with $0\%$ training noise (left panel) and $20\%$ training noise (middle and right panel).
		The figures show the test data containing $5\%$ noise.
		The networks trained with noisy data are more robust to noise than networks trained with noise-less data; contrast the AUC to the corresponding panel of figure~\ref{sec4_noiseTest}. 
	}
	\label{sec4_FHN_noiseTrain}
\end{figure*}

Recall that the results reported in figure~\ref{sec4_noiseTest} correspond to a network trained on noise-free data, and then tested on noisy observational inputs. Interestingly, if we add artificial noise to the training data, the prediction accuracy of all networks increases significantly. 
Figure~\ref{sec4_FHN_noiseTrain} demonstrates the effect of training noise on predicting the quantity of interest $q$.
With $0\%$ training noise, the predicted quantity of interest $\hat q(t)$ deviates drastically from the true value $q(t)$ (left panel of figure~\ref{sec4_FHN_noiseTrain}). 
In contrast, after adding $20\%$ noise to the training data, the predicted quantity of interest closely resembles its true time series (see middle panel of figure~\ref{sec4_FHN_noiseTrain}).
This is also reflected in the AUC as shown in the right panel of figure~\ref{sec4_FHN_noiseTrain} where AUC stays close to 1 as the 
noise intensity in the testing data increases. 
Contrast this with the corresponding figure~\ref{sec4_noiseTest} where no artificial noise was added to the training data.

In summary, adding artificial noise to the training data leads to more accurate extreme event predictions if the real-time observational measurements (i.e., testing data) are noisy. We have made a similar observation for the R\"ossler system and the Kolmogorov flow (not shown here for brevity).
This is in line with previous studies which find that adding noise to the data prevents overfitting~\cite{trainingnoiseref2, trainingnoiseref}.

\subsection{Kolmogorov flow}\label{sec:results_kolm}
In this section, we finally consider the Kolmogorov flow, a particular type of turbulent fluid flow with periodic boundary conditions and sinusoidal forcing. Solving the corresponding Navier--Stokes equations, we 
generate the full simulation data which consists of the velocity field $\vc u(\vc x,t)$. The equations are numerically integrated for $10^5$ time units
and the velocity field is saved on a uniform $128\times 128$
spatial grid at equispaced time intervals $\Delta t=0.2$. 
A detailed description of the Kolmogorov flow is provided in Appendix~\ref{app:kol}.

It is well-known that the energy dissipation rate $D(t)$, i.e., the spatial average of the vorticity field, exhibits extreme events in form of intermittent bursts~\cite{faraz_adjoint} (see figure~\ref{sec4_timeseries}). Therefore, we take the quantity of interest to be the energy dissipation rate, $q = D$.

We consider two sets of neural network inputs for the Kolmogorov flow. 
First, we predict extreme events using the Fourier mode $a(1,0) \in \mathbb{C}$, where $a(\vc k)$ denotes the Fourier mode of the flow corresponding to the wave number $\vc k$. Using a variational method, Farazmand and Sapsis~\cite{Farazmand2017} discovered that the Fourier mode $a(1,0)$ constitute a precursor to extreme events in the Kolmogorov flow. More precisely, the magnitude $|a(1,0)|$ of this mode decreases shortly before a burst in the energy dissipation $D$ is observed. Therefore, we expect that using this mode as the observable will enable accurate prediction of extreme events in this flow. Consequently, we take the first set of observable to be the real and imaginary parts of this Fourier mode, i.e., $p_1 = \Re[a(1,0)]$ and $p_2 = \Im[a(1,0)]$. The results with Fourier observables are discussed in Section~\ref{sec:Kolm_Fourier}.

As a second set of observables, we consider discrete measurements of the vorticity field $\omega(\vc x,t)= \nabla\times \vc u(\vc x,t)$.
More precisely, we consider the vorticity field evaluated on a uniform $3\times 3$ grid at points $\vc x_i \in \{\frac{\pi}{3}, \pi, \frac{5\pi}{3}\} \times \{\frac{\pi}{3}, \pi, \frac{5\pi}{3}\}$. Then the observables are given by $p_i(t)=\omega (\vc x_i,t)$, $i=1,2,\cdots, 9$.
The results with vorticity observables are discussed in Section~\ref{sec:Kolm_vort}.

As we show below, the neural networks trained with Fourier observables vastly outperform those trained with vorticity measurements. 
This is in spite of the fact that Fourier mode data is of smaller dimension (two time series) than the vorticity measurements (nine time series), 
highlighting the importance of the input data for deep learning of extreme events. 
The optimal hyperparameters for training the neural networks are detailed in Appendix~\ref{app:hyperparam_Kolm}.

\subsubsection{Fourier mode}\label{sec:Kolm_Fourier}

As shown in figure~\ref{sec4_summary}, all network structures perform well with the real and imaginary parts of the Fourier mode $a(1,0)$ as input. The performance of the networks are comparable with their AUCs between $0.986$ and $0.988$. Nonetheless, the FF neural network slightly outperforms the LSTM and RC networks. 

The FF network also marginally outperforms the other two networks when performance is measured in terms of NRMSE. 
The NRMSE for the FF network is $0.240$, while the RC network had an error of $0.251$ and the LSTM network had an error of $0.267$. For all networks, the NRMSE varies insignificantly over repeated training, with variations of less than $0.01$. This implies that the learned networks are robust with respect to retraining.

Keeping network structures fixed, we show the effect of increasing the prediction time $\tau$ in figure~\ref{sec4_tau}. 
All network architectures perform similarly as $\tau$ increases, retaining strong predictive power up to $\tau = 3$, where the AUC is approximately $0.9$ for each network.  As $\tau$ increases further above $5$ time units, the predictive power drops off significantly, with the AUC reducing to around $0.5$. 

In figure~\ref{sec4_noiseTest}, we analyze the effect of noisy input data on network performance. 
The LSTM network is remarkably robust to noise, returning AUC values for $20\%$ noise which are comparable to those for noise-free data.
The performance of RC network deteriorates slightly as the noise intensity increases. As in the case of R\"ossler and FHN systems, the FF network performs significantly worse with noisy input data.

In summary, when supplied with the Fourier input data, all network structures produce accurate predictions for moderate prediction times ($\tau< 5$) and moderate noise intensity (less than $10\%$). This should not be surprising since the Fourier mode $a(1,0)$ is known to act as a precursor to extreme events in the Kolmogorov flow~\cite{Farazmand2017}. 
In fact, Farazmand and Sapsis~\cite{Farazmand2017} were able to predict extreme events in this system by tracking the modulus $|a(1,0)|$ without using a deep learning method. 

However, measuring the Fourier mode $a(1,0)$ is not straightforward in practice; to compute the Fourier transform, the flow must be measured on a dense spatial grid.
A more realistic set of observables is velocity or vorticity measurements on a sparse spatial grid. 
As we see in Section~\ref{sec:Kolm_vort}, the prediction accuracy deteriorates significantly if such realistic input data is used for training the neural networks.

\subsubsection{Vorticity samples}\label{sec:Kolm_vort}

Kolmogorov flow with vorticity sampling input data is the most difficult system-data combination we consider. 
As a result, our networks predict extreme events less accurately as compared to other systems.

As shown in figure~\ref{sec4_summary}, when supplied with the vorticity as the input data, the RC network yields the best predictions in terms of maximal AUC and minimal NRMSE. 
Reservoir computing gives an AUC of $0.787$, whereas the LSTM network has an AUC of $0.706.$ 
The feedforward network performs the worst with an AUC of $0.682$.
Note that this is in contrast with previous systems where the FF network outperformed RC when tested on noiseless input data.

Similarly conclusions are drawn when comparing performance besed on NRMSE. RC performs best with a mean NRMSE of $0.552$, relative to $0.597$ for LSTM networks and $0.607$ for FF networks. 
We see that LSTM network exhibits the most variation under multiple training, differing from the mean NRMSE by $0.040$ between different trainings. NRMSE varies significantly less for FF and RC networks, on the order of $10^{-4}$.

Figure~\ref{sec4_tau} shows the prediction performance, measured by AUC, as the prediction time $\tau$ 
increases from $\tau = 0$ to $\tau = 10$. As expected, larger prediction times lead to lower accuracy. The prediction accuracies decline at a similar rate across networks.

As shown in figure~\ref{sec4_noiseTest}, when noise is added to the test data, LSTM and RC networks perform significantly better than the FF network. As in the previous systems, the LSTM network is quite robust to noise, 
with the AUC barely dropping even when the noise intensity increases to $20\%$.
In the Kolmogorov flow with the vorticity input, the RC network is also equally robust to noise.
In contrast, the AUC drops sharply for the FF network as the noise intensity increases.
We find that adding noise to the training data slightly improves the performance of the FF network on noisy test data (not shown here for brevity). Recall that a similar improvement was observed for the FHN system (see figure~\ref{sec4_FHN_noiseTrain}); however, the improvement for the Kolmogorov flow is not as drastic as the
FHN system.

\section{Summary and conclusions} \label{conclusion}
Using deep learning to predict extreme events remains challenging due to the small data problem. 
Even large amounts of observational data contain few extreme events to sufficiently train a neural network.
Here, we investigated whether training on numerically generated data, with sufficient samples from extreme events, 
can overcome this obstacle. In addition, to mimic the practical situation where observable quantities are limited, we did not use the entire simulation data for training and instead used a small subset of the available time series data as network inputs.

To examine this framework, we conducted a thorough study on three dynamical systems (R\"osller system, FitzHugh--Nagumo model, and Kolmogorov flow) and three standard machine learning architectures (Feedfoward, long short-term memory, and reservoir computing networks), resulting in nine total combinations.

\begin{table*}
	\centering
	\caption{The best performing machine learning architecture as measured by the area under the precision-recall curve (or AUC).
		FHN stands for the FitzHugh--Nagumo model and KF stands for the Kolmogorov flow.}
	\begin{tabular}{c|c|c|c|c|}
		\cline{2-5}
		%& \multicolumn{4}{|c|}{Dynamical system} \\
		\cline{2-5}
		& \textbf{R\"ossler} & \textbf{FHN} & \textbf{KF: Fourier} & \textbf{KF: Vorticity} \\ \hline\hline
		\multicolumn{1}{|c||}{\textbf{Noiseless data}} & FF & FF & FF, LSTM, RC & RC\\ \hline
		\multicolumn{1}{|c||}{\textbf{Noisy data}} & LSTM & LSTM & LSTM & RC\\ \hline		
	\end{tabular}\label{tab:bestNet}
\end{table*}

As summarized in Table~\ref{tab:bestNet}, there is no universal answer: no network architecture consistently outperforms
others across different dynamical systems. Nonetheless, some broad conclusions can be drawn from our analysis. For instance, LSTM networks are most robust to observational noise, i.e., they largely maintain their accuracy of extreme event prediction when a moderate amount of noise is added to the observational data even if the training data (obtained from simulations) has no significant noise. In contrast, FF neural networks are most sensitive to observational noise, with their prediction accuracy deteriorating rapidly as the noise intensity increases. 
Note that although RC works best for the Kolmogorov flow with vorticity inputs, its performance is not significantly better than the corresponding LSTM network (see figure~\ref{sec4_noiseTest}).

Furthermore, LSTM networks performed well with minimal fine-tuning of hyperparameters. For LSTM networks, we only had to fine-tune the number of layers and units per layer while for FF and RC networks several hyperparameters had to be fine-tuned to achieve comparable prediction accuracy.

We also find that adding artificial noise to the training data consistently improves the predictions.
Recall that our training data is provided by numerical simulations which contain no significant noise (other than round-off and numerical discretization errors). However, one can add some 
artificial noise to the simulation data to mimic the observational noise. Adding noise to the training data improved the performance of all networks across all systems. This improvement was specifically significant for the FF neural network trained on the FHN data. This observation is in line with previous studies which find that adding noise to the data prevents overfitting~\cite{trainingnoiseref2, trainingnoiseref}.

Another important observation is the sensitivity of the results to the type of observations, i.e., network inputs. This point is clearly demonstrated on the Kolmogorov flow where extreme events are predicted with high accuracy if the input data is obtained from the Fourier mode $a(1,0)$. In contrast, a large number of false positive and false negative predictions are made if the input data consists of sparse vorticity measurements. Using a variational method, Farazmand and Sapsis~\cite{Farazmand2017} had previously found the Fourier mode $a(1,0)$ to play an important role in the formation of extreme events in the Kolmogorov flow. Therefore, when possible, it is advised to discover precursors to extreme events and train the neural networks using these precursors. Discovering precursors to extreme events is itself challenging and remains the subject of ongoing research.~\cite{farazmand2019a} 

{\cb On a related note, assume that a large number of observable time series of the system are available. To maintain a reasonable training cost, one needs to select a subset of the observables as input data. Currently a concrete method to select the optimal subset of the observed time series, ensuring maximal extreme event prediction skill, is missing. More theoretical work in this direction is highly desirable.}

The performance of a deep neural network is sensitive to its hyperparameters, such as number of layers, nodes, and learning rate. Here, we determined the hyperparameters manually using extensive trial-and-error searches.
More systematic methods, such as grid search or random search~\cite{Young2015}, can be used to determine the optimal hyperparameters.

{\cb Finally, the length of training data was here determined in an ad hoc manner. Our future theoretical work
will focus on deriving a lower bound on the number of required extreme events 
in the training data which guarantee accurate prediction of upcoming extreme events in the testing data.}

\section*{Acknowledgments}
This work was supported by the National Science Foundation grant DMS-2051010 and  National Security Agency grants H98230-20-1-0259 and H98230-21-1-0014.

\appendix

\section{Governing equations} \label{appendix_equations}
\subsection{R\"ossler system}
\label{app:rossler}
The R\"ossler system is defined by the ODEs,

\begin{align}
	\begin{split}
		\dot{x}_1 &= -x_2-x_3, \\
		\dot{x}_2 &= x_1+ax_2, \\
		\dot{x}_3 &= b+x_3(x_1-c),
	\end{split}
\end{align}
with $a=0.2$, $b=0.2$, and $c=5.7$. 
This system exhibits a chaotic attractor and intermittent extreme event \emph{bursts} in the $x_3$ component (see figure~\ref{sec4_timeseries}). 
Therefore, we consider the quantity of interest, 
\begin{equation}
	q(\vc x(t)) = x_3(t)
\end{equation} 
and the partial observations $p_1(\vc x(t)) =x_1(t),\, p_2(\vc x(t)) =x_2(t)$.
The extreme event threshold is $q_e = 10$.

To generate trajectory data, we perform numerical integration using the Runge-Kutta method, as implemented in MATLAB's \texttt{ode45}. 
We set the initial condition as $(0,1,0.1)$ and integrate the system for $500$ time units. The results are saved every $\Delta t = 0.05$ time units. 

\subsection{FitzHugh--Nagumo system}
\label{app:FHN}

The FitzHugh--Nagumo system consists of $n$ excitable units $(v_i,w_i)$, $i = 1,\ldots,n$. These pairs can be interpreted as neurons, where $v_i$ is the voltage and $w_i$ is a recovery term. The dynamics are described by a system of $2n$ differential equations,
\begin{align}
	\begin{split}
		\dot{v}_i &= v_i (a_i - v_i)(v_i - 1) - w_i + k \sum_{j=1}^{n}A_{ij}(v_j - v_i),\\
		\dot{w}_i &= b_i v_i - c_i w_i.
	\end{split}
\end{align}
The full state is described by $\vc x(t) = [v_1(t), w_1(t),..., v_n(t), w_n(t)].$
Parameters $a_i$, $b_i$, and $c_i$ describe the dynamics of each unit internally, $k$ is the coupling strength among units, and $A_{ij}$ describes the coupling connections, where $A_{ij} = A_{ji} = 1$ indicates units $i$ and $j$ are coupled.
We choose the parameters in case B of Ref.~\onlinecite{FHNparameters}, which shows chaotic behavior with extreme events.
This system has $n=101$ completely coupled units, meaning $A_{ij} = 1, \forall i,j$.
The parameters are $a_i = -0.02651$ and $c_i = 0.02$ for all units while $b_i = 0.006 + 0.008 (i-1)/(n-1)$.
These choices, especially the coupling strength $k=0.00128$, are further discussed by Feudel et. al.~\cite{feudel2014}.

The quantity of interest is the average voltage of all the nodes, 
\begin{equation}
	q(\vc x(t)) = \frac{1}{n}\sum_{i=1}^n v_i(t).
\end{equation}
The units of FHN oscillator occasionally synchronize, leading to intermittent bursts in the average $q$. 
We choose an  extreme event threshold $q_e = 0.3$. We use $p_1(\vc x(t)) = v_1(t)$, $p_2(\vc x(t)) = w_1(t)$ as observables, but other units give comparable performance.

% XXX
We use a Runge--Kutta method in MATLAB's \texttt{ode45} to numerically integrate the FHN system for $2\times 10^5$ time units and save the results every $\Delta t = 1$ time units. We discard the first 100 time units as transient data from the initial condition $(v_i, w_i) = (0.1, 0.1)$, $i=1,\ldots, 101$.

\subsection{Kolmogorov flow}
\label{app:kol}
We consider the Navier-Stokes equations for incompressible fluid flow,
\begin{equation}
	\begin{gathered}
		\partial_t \mathbf{u} + \mathbf{u} \cdot \grad \mathbf{u}=  - \grad p + \nu \Delta\mathbf{u} +\vc F\\
		\grad \cdot \mathbf{u} = 0, 
	\end{gathered}
\end{equation}
where $\mathbf{u}(\vc x,t)$ is the velocity field, $p(\vc x,t)$ is the pressure field, $\nu$ is the kinematic viscosity, and $\vc F$ is the external forcing function. 
We specify the Kolmogorov flow as a specific two-dimensional realization of the Navier-Stokes equations. 
The domain is a doubly-periodic box of size $2\pi\times 2\pi$.
A turbulent regime is chosen by using the Reynolds number $Re = \nu^{-1} = 40$. 
The external forcing function is the sinusoidal shearing function, $\vc F(x,y) = \sin(4y)\vc e_1$, where $\vc e_1$ is the standard basis vector in the $x$ direction in $\R^2$.
We numerically integrate the Kolmogorov flow using a Fourier pseuedospectral method in space with $2^{7}$ modes in each dimension and a fourth-order Runge–Kutta scheme
for time stepping. The flow is evolved for $10^5$ time units and results are saved every $\Delta t = 0.2$ time units. We start the simulation from a random initial condition and discard the first $20$ time units of data as transients. 

Here, the quantity of interest is the energy dissipation rate,
\begin{align}
	q(\mathbf{u}) &= \frac{\nu}{| \Omega |} \int_\Omega | \grad \mathbf{u} |^2 \, \id\vc x,
\end{align}
where $| \Omega |=(2\pi)^2$ is the area of the spatial domain $\Omega = [0,2\pi]\times[0,2\pi]$. It is known that the Kolmogorov flow exhibits intermittent bursts where
the energy dissipation rate $q$ increases to several standard deviations above its mean value~\cite{faraz_adjoint} (see figure~\ref{sec4_timeseries}).
We use the extreme event threshold $q_e=0.194$ which coincides with the mean plus twice the standard deviation of energy dissipation rate.

It has been recently discovered that these extreme events are instigated by the energy transfer between particular Fourier modes~\cite{Farazmand2017}. Consider the Fourier series expansion of the velocity field,
\begin{equation}
	\vc u(\vc x,t) = \sum_{\vc k\in\mathbb Z^2}\frac{a(\vc k,t)}{|\vc k|^2}
	\begin{pmatrix}
		k_y\\
		-k_x
	\end{pmatrix}e^{\hat i\vc k\cdot\vc x},
\end{equation}
where $\vc k=(k_x,k_y)$ denotes the wave number and $a(\vc k,t)\in \mathbb C$ denotes the corresponding Fourier coefficient.
Note that the incompressibility of the velocity field ($\nabla\cdot \vc u=0$) is used to reduce the Fourier coefficients to scalars. Since the velocity is real-valued, we have
$a(-\vc k,t)=-a(\vc k,t)^\ast$.

Using a variational method, Farazmand and Sapsis~\cite{Farazmand2017} showed that the extreme energy dissipation events are preceeded by an energy transfer from mode $a(1,0,t)$
to the mode $a(0,4,t)$. Motivated by this work, we use the real and imaginary parts of $a(1,0,t)$ as the first set of observables for the Kolmogorov flow. 
More precisely, the Fourier observables are $p_1(t) = \Re[a(1,0,t)]$ and $p_2(t) = \Im[a(1,0,t)]$.

A second set of observables are obtained through the vorticity field $\omega =(\nabla\times \vc u)\cdot \vc e_3$, where $\vc e_3$ is the unit vector normal to the planar fluid domain.
We evaluate the vorticity field at $9$ distinct spatial locations $\vc x_i \in \{\frac{\pi}{3}, \pi, \frac{5\pi}{3}\} \times \{\frac{\pi}{3}, \pi, \frac{5\pi}{3}\}$, leading to $9$ observables 
$p_i(t) = \omega (\vc x_i,t)$.

\section{Network architectures} \label{netarchapp}
In this section, we discuss the optimal parameters found for each of our example systems and the neural networks. 
The hyperparameters are found manually by trial-and-error. Therefore, the word `optimal' is used loosely; it only refers to the optimal combination of hyperparameters we
investigated, not optimal among all possible combinations.
All networks were optimised on noise-free data before exploring network resilience to noise in training and testing data. 
All hyperparameters not mentioned here were set to the default values (Matlab default values for FF and LSTM networks, and \texttt{easyesn} library for RC).

\subsection{R\"ossler system}
\label{app:rossler_param}
% XXX
The best performing FF network architecture for the R\"ossler system consists of 3 layers with 6 nodes each with a $\tanh$ activation function. The architecture used $m=3$ time delays, $s=1$ time unit apart.
These time delays improve the AUC and NRMSE of the network.

The optimal LSTM architecture consists of 2 layers with 55 nodes per layer.
The optimal RC network contains $850$ nodes in the reservoir. 
However with $5\%$ testing noise the network AUC quickly decreased to $0.0359$, suggesting that the network is overfitted. 
Re-optimizing the RC network with noisy data yields a drastically different optimal network: $50$ nodes in the reservoir.
Retraining the network on noiseless data with $50$ nodes shows that at $0\%$ testing noise the AUC is $0.9811$, but the predictive power remains reasonable for higher levels of noise.
At $5\%$, the AUC decreases to $0.9623$.
The optimal spectral radius is $\rho = 0.3$ and leaking rate is $\ell=1$.
The optimal input density is 1, the reservoir density is $0.2$ and the regularization weight is $10^{-4}$.

\subsection{FitzHugh--Nagumo system}
\label{app:FHN_param}
The optimal FF network architecture is 3 layers with 8 nodes, each with the $\tanh$ activation function. The architecture used $m=2$ time delays with $s=2$ time unit.
% when the data is augmented with 2 time delays, at 2 and 4 time units prior.
However, this architecture is susceptible to noise in the testing data.
We address the overfitting by adding noise to the training data.
When varying the magnitude of noise in the training data, we find that increasing noise in the training data notably increases performance with high noise but slightly decreases performance with low noise.
We choose to favor robustness and add relatively high 20\% noise to the training data.
Other alterations to neural network training such as reducing the number of epochs trained and adding a regularization term that penalized large weights did not produce significant improvements.

The LSTM network architecture of 2 layers with 64 hidden units each was found to be optimal.
The time it takes to train the LSTM networks is much longer than FF and RC networks, as expected.

We optimized a wider range of parameters for the RC networks due to poor performance with default parameters.
The optimal architecture used 500 nodes with spectral radius $\rho =0.9$ and leaking rate $\ell = 0.3$.
The input density is 1  with reservoir density $0.9$ and regularization weight $10^{-4}.$
The spectral radius had the most significant effect on performance in the parameter ranges we examined.

% The optimal spectral radius used $\rho = 0.3$ and leaking rate $lr=1$. The optimal input density is 1 while the reservoir density is $0.2$ with regularization weight $10^{-4}$.
\subsection{Kolmogorov flow}\label{app:hyperparam_Kolm}
\subsubsection{Fourier mode}
\label{app:fourier_param}
The best performing FF network architecture uses 3 layers, each with 4 nodes and the $\tanh$ activation function. 
We augment input data with $m=8$ time delays, taken at time steps of $s=0.2$ time units. 
The best performing LSTM network has 3 layers, each with 32 nodes. 

The best performing RC network has $600$ nodes and a spectral radius of $\rho = 0.9$ with leaking rate $\ell =1$. 
The network also uses an input density of $1$, reservoir density of $\rho = 0.2$, and regularization weight $0.1.$

\subsubsection{Vorticity samples}
\label{app:vorticity_param}
The best performing FF neural network has $4$ layers with $8$ nodes per layer, the $\tanh$ activation function, and $m=12$ time delays each $s=0.2$ time units apart. 
The optimal FF network taking in vorticity samples requires more layers and nodes because the vorticity data is higher dimensional than the Fourier modes. 

The LSTM network with 2 layers and 16 nodes per layer performed best.
Using more than 16 nodes per layer increases computational cost without improving the predictions. 
In fact, networks with $2$ LSTM layers of $64$ or more nodes per layer exhibit signs of overfitting. 
We added dropout layers to try to fix this issue, but prediction accuracy, as measured by NRMSE and AUC, did not improve when measured on testing data. 

The best RC network has 1,000 nodes, spectral radius $\rho =0.2$, and leaking rate  $\ell = 0.3$.
The optimal input density is $0.3$ with reservoir density of $0.2$ and regularization weight $10^{-4}$.

\section*{Data availability}
The data and code that support the findings of this study are available
at \url{https://github.com/mfarazmand/DeepLearningExtremeEvents}

%%\bibliographystyle{unsrt}
%\bibliography{bibliog}

\begin{thebibliography}{46}%
	\makeatletter
	\providecommand \@ifxundefined [1]{%
		\@ifx{#1\undefined}
	}%
	\providecommand \@ifnum [1]{%
		\ifnum #1\expandafter \@firstoftwo
		\else \expandafter \@secondoftwo
		\fi
	}%
	\providecommand \@ifx [1]{%
		\ifx #1\expandafter \@firstoftwo
		\else \expandafter \@secondoftwo
		\fi
	}%
	\providecommand \natexlab [1]{#1}%
	\providecommand \enquote  [1]{``#1''}%
	\providecommand \bibnamefont  [1]{#1}%
	\providecommand \bibfnamefont [1]{#1}%
	\providecommand \citenamefont [1]{#1}%
	\providecommand \href@noop [0]{\@secondoftwo}%
	\providecommand \href [0]{\begingroup \@sanitize@url \@href}%
	\providecommand \@href[1]{\@@startlink{#1}\@@href}%
	\providecommand \@@href[1]{\endgroup#1\@@endlink}%
	\providecommand \@sanitize@url [0]{\catcode `\\12\catcode `\$12\catcode
		`\&12\catcode `\#12\catcode `\^12\catcode `\_12\catcode `\%12\relax}%
	\providecommand \@@startlink[1]{}%
	\providecommand \@@endlink[0]{}%
	\providecommand \url  [0]{\begingroup\@sanitize@url \@url }%
	\providecommand \@url [1]{\endgroup\@href {#1}{\urlprefix }}%
	\providecommand \urlprefix  [0]{URL }%
	\providecommand \Eprint [0]{\href }%
	\providecommand \doibase [0]{http://dx.doi.org/}%
	\providecommand \selectlanguage [0]{\@gobble}%
	\providecommand \bibinfo  [0]{\@secondoftwo}%
	\providecommand \bibfield  [0]{\@secondoftwo}%
	\providecommand \translation [1]{[#1]}%
	\providecommand \BibitemOpen [0]{}%
	\providecommand \bibitemStop [0]{}%
	\providecommand \bibitemNoStop [0]{.\EOS\space}%
	\providecommand \EOS [0]{\spacefactor3000\relax}%
	\providecommand \BibitemShut  [1]{\csname bibitem#1\endcsname}%
	\let\auto@bib@innerbib\@empty
	%</preamble>
	\bibitem [{\citenamefont {Comfort}, \citenamefont {Boin},\ and\ \citenamefont
		{Demchak}(2010)}]{comfort2010}%
	\BibitemOpen
	\bibfield  {author} {\bibinfo {author} {\bibfnamefont {L.~K.}\ \bibnamefont
			{Comfort}}, \bibinfo {author} {\bibfnamefont {A.}~\bibnamefont {Boin}}, \
		and\ \bibinfo {author} {\bibfnamefont {C.~C.}\ \bibnamefont {Demchak}},\
	}\href@noop {} {\emph {\bibinfo {title} {Designing resilience: Preparing for
				extreme events}}}\ (\bibinfo  {publisher} {University of Pittsburgh},\
	\bibinfo {year} {2010})\BibitemShut {NoStop}%
	\bibitem [{\citenamefont {Farazmand}\ and\ \citenamefont
		{Sapsis}(2019)}]{farazmand2019a}%
	\BibitemOpen
	\bibfield  {author} {\bibinfo {author} {\bibfnamefont {M.}~\bibnamefont
			{Farazmand}}\ and\ \bibinfo {author} {\bibfnamefont {T.~P.}\ \bibnamefont
			{Sapsis}},\ }\bibfield  {title} {\enquote {\bibinfo {title} {Extreme events:
				{M}echanisms and {P}rediction},}\ }\href {\doibase 10.1115/1.4042065}
	{\bibfield  {journal} {\bibinfo  {journal} {Applied Mechanics Review}\
		}\textbf {\bibinfo {volume} {71}} (\bibinfo {year} {2019}),\
		10.1115/1.4042065}\BibitemShut {NoStop}%
	\bibitem [{\citenamefont {McPhillips}\ \emph {et~al.}(2018)\citenamefont
		{McPhillips}, \citenamefont {Chang}, \citenamefont {Chester}, \citenamefont
		{Depietri}, \citenamefont {Friedman}, \citenamefont {Grimm}, \citenamefont
		{Kominoski}, \citenamefont {McPhearson}, \citenamefont {Méndez-Lázaro},
		\citenamefont {Rosi},\ and\ \citenamefont {Shafiei~Shiva}}]{Shafiei18}%
	\BibitemOpen
	\bibfield  {author} {\bibinfo {author} {\bibfnamefont {L.~E.}\ \bibnamefont
			{McPhillips}}, \bibinfo {author} {\bibfnamefont {H.}~\bibnamefont {Chang}},
		\bibinfo {author} {\bibfnamefont {M.~V.}\ \bibnamefont {Chester}}, \bibinfo
		{author} {\bibfnamefont {Y.}~\bibnamefont {Depietri}}, \bibinfo {author}
		{\bibfnamefont {E.}~\bibnamefont {Friedman}}, \bibinfo {author}
		{\bibfnamefont {N.~B.}\ \bibnamefont {Grimm}}, \bibinfo {author}
		{\bibfnamefont {J.~S.}\ \bibnamefont {Kominoski}}, \bibinfo {author}
		{\bibfnamefont {T.}~\bibnamefont {McPhearson}}, \bibinfo {author}
		{\bibfnamefont {P.}~\bibnamefont {Méndez-Lázaro}}, \bibinfo {author}
		{\bibfnamefont {E.~J.}\ \bibnamefont {Rosi}}, \ and\ \bibinfo {author}
		{\bibfnamefont {J.}~\bibnamefont {Shafiei~Shiva}},\ }\bibfield  {title}
	{\enquote {\bibinfo {title} {Defining extreme events: {A} cross-disciplinary
				review},}\ }\href {\doibase https://doi.org/10.1002/2017EF000686} {\bibfield
		{journal} {\bibinfo  {journal} {Earth's Future}\ }\textbf {\bibinfo {volume}
			{6}},\ \bibinfo {pages} {441--455} (\bibinfo {year} {2018})}\BibitemShut
	{NoStop}%
	\bibitem [{\citenamefont {Callaham}, \citenamefont {Maeda},\ and\ \citenamefont
		{Brunton}(2019)}]{Callaham2019}%
	\BibitemOpen
	\bibfield  {author} {\bibinfo {author} {\bibfnamefont {J.~L.}\ \bibnamefont
			{Callaham}}, \bibinfo {author} {\bibfnamefont {K.}~\bibnamefont {Maeda}}, \
		and\ \bibinfo {author} {\bibfnamefont {S.~L.}\ \bibnamefont {Brunton}},\
	}\bibfield  {title} {\enquote {\bibinfo {title} {Robust flow reconstruction
				from limited measurements via sparse representation},}\ }\href {\doibase
		10.1103/PhysRevFluids.4.103907} {\bibfield  {journal} {\bibinfo  {journal}
			{Phys. Rev. Fluids}\ }\textbf {\bibinfo {volume} {4}},\ \bibinfo {pages}
		{103907} (\bibinfo {year} {2019})}\BibitemShut {NoStop}%
	\bibitem [{\citenamefont {Chu}\ and\ \citenamefont
		{Farazmand}(2021)}]{Chu2021}%
	\BibitemOpen
	\bibfield  {author} {\bibinfo {author} {\bibfnamefont {B.}~\bibnamefont
			{Chu}}\ and\ \bibinfo {author} {\bibfnamefont {M.}~\bibnamefont
			{Farazmand}},\ }\bibfield  {title} {\enquote {\bibinfo {title} {Data-driven
				prediction of multistable systems from sparse measurements},}\ }\href
	{\doibase 10.1063/5.0046203} {\bibfield  {journal} {\bibinfo  {journal}
			{Chaos}\ }\textbf {\bibinfo {volume} {31}},\ \bibinfo {pages} {063118}
		(\bibinfo {year} {2021})}\BibitemShut {NoStop}%
	\bibitem [{\citenamefont {Kramer}\ \emph {et~al.}(2017)\citenamefont {Kramer},
		\citenamefont {Grover}, \citenamefont {Boufounos}, \citenamefont {Nabi},\
		and\ \citenamefont {Benosman}}]{Kramer2017}%
	\BibitemOpen
	\bibfield  {author} {\bibinfo {author} {\bibfnamefont {B.}~\bibnamefont
			{Kramer}}, \bibinfo {author} {\bibfnamefont {P.}~\bibnamefont {Grover}},
		\bibinfo {author} {\bibfnamefont {P.}~\bibnamefont {Boufounos}}, \bibinfo
		{author} {\bibfnamefont {S.}~\bibnamefont {Nabi}}, \ and\ \bibinfo {author}
		{\bibfnamefont {M.}~\bibnamefont {Benosman}},\ }\bibfield  {title} {\enquote
		{\bibinfo {title} {Sparse sensing and {DMD}-based identification of flow
				regimes and bifurcations in complex flows},}\ }\href {\doibase
		10.1137/15M104565X} {\bibfield  {journal} {\bibinfo  {journal} {SIAM Journal
				on Applied Dynamical Systems}\ }\textbf {\bibinfo {volume} {16}},\ \bibinfo
		{pages} {1164--1196} (\bibinfo {year} {2017})}\BibitemShut {NoStop}%
	\bibitem [{\citenamefont {Farazmand}\ and\ \citenamefont
		{Sapsis}(2017{\natexlab{a}})}]{Farazmand2017b}%
	\BibitemOpen
	\bibfield  {author} {\bibinfo {author} {\bibfnamefont {M.}~\bibnamefont
			{Farazmand}}\ and\ \bibinfo {author} {\bibfnamefont {T.~P.}\ \bibnamefont
			{Sapsis}},\ }\bibfield  {title} {\enquote {\bibinfo {title} {Reduced-order
				prediction of rogue waves in two-dimensional deep-water waves},}\ }\href@noop
	{} {\bibfield  {journal} {\bibinfo  {journal} {J. Comput. Phys.}\ }\textbf
		{\bibinfo {volume} {340}},\ \bibinfo {pages} {418 -- 434} (\bibinfo {year}
		{2017}{\natexlab{a}})}\BibitemShut {NoStop}%
	\bibitem [{\citenamefont {Mendez}\ and\ \citenamefont
		{Farazmand}(2021)}]{mendez2020}%
	\BibitemOpen
	\bibfield  {author} {\bibinfo {author} {\bibfnamefont {A.}~\bibnamefont
			{Mendez}}\ and\ \bibinfo {author} {\bibfnamefont {M.}~\bibnamefont
			{Farazmand}},\ }\bibfield  {title} {\enquote {\bibinfo {title} {Investigating
				climate tipping points under various emission reduction and carbon capture
				scenarios with a stochastic climate model},}\ }\href@noop {} {\bibfield
		{journal} {\bibinfo  {journal} {Proc. Royal Soc. A, In press,
				arXiv:2012.01613}\ } (\bibinfo {year} {2021})}\BibitemShut {NoStop}%
	\bibitem [{\citenamefont {Bucklew}(2004)}]{bucklew2004}%
	\BibitemOpen
	\bibfield  {author} {\bibinfo {author} {\bibfnamefont {J.}~\bibnamefont
			{Bucklew}},\ }\href@noop {} {\emph {\bibinfo {title} {Introduction to rare
				event simulation}}}\ (\bibinfo  {publisher} {Springer Science \& Business
		Media},\ \bibinfo {year} {2004})\BibitemShut {NoStop}%
	\bibitem [{\citenamefont {Dematteis}, \citenamefont {Grafke},\ and\
		\citenamefont {Vanden-Eijnden}(2018)}]{Dematteis2018}%
	\BibitemOpen
	\bibfield  {author} {\bibinfo {author} {\bibfnamefont {G.}~\bibnamefont
			{Dematteis}}, \bibinfo {author} {\bibfnamefont {T.}~\bibnamefont {Grafke}}, \
		and\ \bibinfo {author} {\bibfnamefont {E.}~\bibnamefont {Vanden-Eijnden}},\
	}\bibfield  {title} {\enquote {\bibinfo {title} {Rogue waves and large
				deviations in deep sea},}\ }\href {\doibase 10.1073/pnas.1710670115}
	{\bibfield  {journal} {\bibinfo  {journal} {Proc. Natl. Acad. Sci.}\ }\textbf
		{\bibinfo {volume} {115}},\ \bibinfo {pages} {855--860} (\bibinfo {year}
		{2018})}\BibitemShut {NoStop}%
	\bibitem [{\citenamefont {Mohamad}\ and\ \citenamefont
		{Sapsis}(2018)}]{Mohamad2018}%
	\BibitemOpen
	\bibfield  {author} {\bibinfo {author} {\bibfnamefont {M.~A.}\ \bibnamefont
			{Mohamad}}\ and\ \bibinfo {author} {\bibfnamefont {T.~P.}\ \bibnamefont
			{Sapsis}},\ }\bibfield  {title} {\enquote {\bibinfo {title} {Sequential
				sampling strategy for extreme event statistics in nonlinear dynamical
				systems},}\ }\href {\doibase 10.1073/pnas.1813263115} {\bibfield  {journal}
		{\bibinfo  {journal} {Proceedings of the National Academy of Sciences}\
		}\textbf {\bibinfo {volume} {115}},\ \bibinfo {pages} {11138--11143}
		(\bibinfo {year} {2018})}\BibitemShut {NoStop}%
	\bibitem [{\citenamefont {Brunton}, \citenamefont {Noack},\ and\ \citenamefont
		{Koumoutsakos}(2020)}]{brunton2020}%
	\BibitemOpen
	\bibfield  {author} {\bibinfo {author} {\bibfnamefont {S.~L.}\ \bibnamefont
			{Brunton}}, \bibinfo {author} {\bibfnamefont {B.~R.}\ \bibnamefont {Noack}},
		\ and\ \bibinfo {author} {\bibfnamefont {P.}~\bibnamefont {Koumoutsakos}},\
	}\bibfield  {title} {\enquote {\bibinfo {title} {Machine learning for fluid
				mechanics},}\ }\href {\doibase 10.1146/annurev-fluid-010719-060214}
	{\bibfield  {journal} {\bibinfo  {journal} {Annual Review of Fluid
				Mechanics}\ }\textbf {\bibinfo {volume} {52}},\ \bibinfo {pages} {477--508}
		(\bibinfo {year} {2020})},\ \Eprint
	{http://arxiv.org/abs/https://doi.org/10.1146/annurev-fluid-010719-060214}
	{https://doi.org/10.1146/annurev-fluid-010719-060214} \BibitemShut {NoStop}%
	\bibitem [{\citenamefont {Fawaz}\ \emph {et~al.}(2019)\citenamefont {Fawaz},
		\citenamefont {Forestier}, \citenamefont {Weber}, \citenamefont {Idoumghar},\
		and\ \citenamefont {Muller}}]{fawaz2019}%
	\BibitemOpen
	\bibfield  {author} {\bibinfo {author} {\bibfnamefont {H.~I.}\ \bibnamefont
			{Fawaz}}, \bibinfo {author} {\bibfnamefont {G.}~\bibnamefont {Forestier}},
		\bibinfo {author} {\bibfnamefont {J.}~\bibnamefont {Weber}}, \bibinfo
		{author} {\bibfnamefont {L.}~\bibnamefont {Idoumghar}}, \ and\ \bibinfo
		{author} {\bibfnamefont {P.}~\bibnamefont {Muller}},\ }\bibfield  {title}
	{\enquote {\bibinfo {title} {Deep learning for time series classification: a
				review},}\ }\href {\doibase 10.1007/s10618-019-00619-1} {\bibfield  {journal}
		{\bibinfo  {journal} {Data mining and knowledge discovery}\ }\textbf
		{\bibinfo {volume} {33}},\ \bibinfo {pages} {917--963} (\bibinfo {year}
		{2019})}\BibitemShut {NoStop}%
	\bibitem [{\citenamefont {Karniadakis}\ \emph {et~al.}(2021)\citenamefont
		{Karniadakis}, \citenamefont {Kevrekidis}, \citenamefont {Lu}, \citenamefont
		{Perdikaris}, \citenamefont {Wang},\ and\ \citenamefont
		{Yang}}]{karniadakis2021}%
	\BibitemOpen
	\bibfield  {author} {\bibinfo {author} {\bibfnamefont {G.~E.}\ \bibnamefont
			{Karniadakis}}, \bibinfo {author} {\bibfnamefont {I.~G.}\ \bibnamefont
			{Kevrekidis}}, \bibinfo {author} {\bibfnamefont {L.}~\bibnamefont {Lu}},
		\bibinfo {author} {\bibfnamefont {P.}~\bibnamefont {Perdikaris}}, \bibinfo
		{author} {\bibfnamefont {S.}~\bibnamefont {Wang}}, \ and\ \bibinfo {author}
		{\bibfnamefont {L.}~\bibnamefont {Yang}},\ }\bibfield  {title} {\enquote
		{\bibinfo {title} {Physics-informed machine learning},}\ }\href@noop {}
	{\bibfield  {journal} {\bibinfo  {journal} {Nature Reviews Physics}\ }\textbf
		{\bibinfo {volume} {3}},\ \bibinfo {pages} {422--440} (\bibinfo {year}
		{2021})}\BibitemShut {NoStop}%
	\bibitem [{\citenamefont {Kochkov}\ \emph {et~al.}(2021)\citenamefont
		{Kochkov}, \citenamefont {Smith}, \citenamefont {Alieva}, \citenamefont
		{Wang}, \citenamefont {Brenner},\ and\ \citenamefont {Hoyer}}]{brenner2021}%
	\BibitemOpen
	\bibfield  {author} {\bibinfo {author} {\bibfnamefont {D.}~\bibnamefont
			{Kochkov}}, \bibinfo {author} {\bibfnamefont {J.~A.}\ \bibnamefont {Smith}},
		\bibinfo {author} {\bibfnamefont {A.}~\bibnamefont {Alieva}}, \bibinfo
		{author} {\bibfnamefont {Q.}~\bibnamefont {Wang}}, \bibinfo {author}
		{\bibfnamefont {M.~P.}\ \bibnamefont {Brenner}}, \ and\ \bibinfo {author}
		{\bibfnamefont {S.}~\bibnamefont {Hoyer}},\ }\bibfield  {title} {\enquote
		{\bibinfo {title} {Machine learning{\textendash}accelerated computational
				fluid dynamics},}\ }\href {\doibase 10.1073/pnas.2101784118} {\bibfield
		{journal} {\bibinfo  {journal} {Proceedings of the National Academy of
				Sciences}\ }\textbf {\bibinfo {volume} {118}} (\bibinfo {year} {2021}),\
		10.1073/pnas.2101784118},\ \Eprint
	{http://arxiv.org/abs/https://www.pnas.org/content/118/21/e2101784118.full.pdf}
	{https://www.pnas.org/content/118/21/e2101784118.full.pdf} \BibitemShut
	{NoStop}%
	\bibitem [{\citenamefont {Pathak}\ \emph {et~al.}(2017)\citenamefont {Pathak},
		\citenamefont {Lu}, \citenamefont {Hunt}, \citenamefont {Girvan},\ and\
		\citenamefont {Ott}}]{Pathak2017}%
	\BibitemOpen
	\bibfield  {author} {\bibinfo {author} {\bibfnamefont {J.}~\bibnamefont
			{Pathak}}, \bibinfo {author} {\bibfnamefont {Z.}~\bibnamefont {Lu}}, \bibinfo
		{author} {\bibfnamefont {B.~R.}\ \bibnamefont {Hunt}}, \bibinfo {author}
		{\bibfnamefont {M.}~\bibnamefont {Girvan}}, \ and\ \bibinfo {author}
		{\bibfnamefont {E.}~\bibnamefont {Ott}},\ }\bibfield  {title} {\enquote
		{\bibinfo {title} {Using machine learning to replicate chaotic attractors and
				calculate lyapunov exponents from data},}\ }\href {\doibase
		10.1063/1.5010300} {\bibfield  {journal} {\bibinfo  {journal} {Chaos: An
				Interdisciplinary Journal of Nonlinear Science}\ }\textbf {\bibinfo {volume}
			{27}},\ \bibinfo {pages} {121102} (\bibinfo {year} {2017})},\ \Eprint
	{http://arxiv.org/abs/https://doi.org/10.1063/1.5010300}
	{https://doi.org/10.1063/1.5010300} \BibitemShut {NoStop}%
	\bibitem [{\citenamefont {Ding}\ \emph {et~al.}(2019)\citenamefont {Ding},
		\citenamefont {Zhang}, \citenamefont {Pan}, \citenamefont {Yang},\ and\
		\citenamefont {He}}]{ding2019}%
	\BibitemOpen
	\bibfield  {author} {\bibinfo {author} {\bibfnamefont {D.}~\bibnamefont
			{Ding}}, \bibinfo {author} {\bibfnamefont {M.}~\bibnamefont {Zhang}},
		\bibinfo {author} {\bibfnamefont {X.}~\bibnamefont {Pan}}, \bibinfo {author}
		{\bibfnamefont {M.}~\bibnamefont {Yang}}, \ and\ \bibinfo {author}
		{\bibfnamefont {X.}~\bibnamefont {He}},\ }\bibfield  {title} {\enquote
		{\bibinfo {title} {Modeling extreme events in time series prediction},}\ }in\
	\href {\doibase 10.1145/3292500.3330896} {\emph {\bibinfo {booktitle}
			{Proceedings of the 25th ACM SIGKDD International Conference on Knowledge
				Discovery \& Data Mining}}},\ \bibinfo {series and number} {KDD '19}\
	(\bibinfo  {publisher} {Association for Computing Machinery},\ \bibinfo
	{address} {New York, NY, USA},\ \bibinfo {year} {2019})\ p.\ \bibinfo {pages}
	{1114–1122}\BibitemShut {NoStop}%
	\bibitem [{\citenamefont {Qi}\ and\ \citenamefont {Majda}(2020)}]{majda2020}%
	\BibitemOpen
	\bibfield  {author} {\bibinfo {author} {\bibfnamefont {D.}~\bibnamefont
			{Qi}}\ and\ \bibinfo {author} {\bibfnamefont {A.~J.}\ \bibnamefont {Majda}},\
	}\bibfield  {title} {\enquote {\bibinfo {title} {Using machine learning to
				predict extreme events in complex systems},}\ }\href {\doibase
		10.1073/pnas.1917285117} {\bibfield  {journal} {\bibinfo  {journal}
			{Proceedings of the National Academy of Sciences}\ }\textbf {\bibinfo
			{volume} {117}},\ \bibinfo {pages} {52--59} (\bibinfo {year} {2020})},\
	\Eprint {http://arxiv.org/abs/https://www.pnas.org/content/117/1/52.full.pdf}
	{https://www.pnas.org/content/117/1/52.full.pdf} \BibitemShut {NoStop}%
	\bibitem [{\citenamefont {Rudy}\ and\ \citenamefont
		{Sapsis}(2021{\natexlab{a}})}]{Rudy2021}%
	\BibitemOpen
	\bibfield  {author} {\bibinfo {author} {\bibfnamefont {S.~H.}\ \bibnamefont
			{Rudy}}\ and\ \bibinfo {author} {\bibfnamefont {T.~P.}\ \bibnamefont
			{Sapsis}},\ }\bibfield  {title} {\enquote {\bibinfo {title} {Output-weighted
				and relative entropy loss functions for deep learning precursors of extreme
				events},}\ }\href@noop {} {\bibfield  {journal} {\bibinfo  {journal} {CoRR}\
		}\textbf {\bibinfo {volume} {abs/2112.00825}} (\bibinfo {year}
		{2021}{\natexlab{a}})},\ \Eprint {http://arxiv.org/abs/2112.00825}
	{2112.00825} \BibitemShut {NoStop}%
	\bibitem [{\citenamefont {Meiyazhagan}, \citenamefont {Sudharsan},\ and\
		\citenamefont {Senthilvelan}(2021)}]{senthilvelan2021}%
	\BibitemOpen
	\bibfield  {author} {\bibinfo {author} {\bibfnamefont {J.}~\bibnamefont
			{Meiyazhagan}}, \bibinfo {author} {\bibfnamefont {S.}~\bibnamefont
			{Sudharsan}}, \ and\ \bibinfo {author} {\bibfnamefont {M.}~\bibnamefont
			{Senthilvelan}},\ }\bibfield  {title} {\enquote {\bibinfo {title} {Model-free
				prediction of emergence of extreme events in a parametrically driven
				nonlinear dynamical system by deep learning},}\ }\href@noop {} {\bibfield
		{journal} {\bibinfo  {journal} {The European Physical Journal B}\ }\textbf
		{\bibinfo {volume} {94}},\ \bibinfo {pages} {1--13} (\bibinfo {year}
		{2021})}\BibitemShut {NoStop}%
	\bibitem [{\citenamefont {Lellep}\ \emph {et~al.}(2020)\citenamefont {Lellep},
		\citenamefont {Prexl}, \citenamefont {Linkmann},\ and\ \citenamefont
		{Eckhardt}}]{lellep2020}%
	\BibitemOpen
	\bibfield  {author} {\bibinfo {author} {\bibfnamefont {M.}~\bibnamefont
			{Lellep}}, \bibinfo {author} {\bibfnamefont {J.}~\bibnamefont {Prexl}},
		\bibinfo {author} {\bibfnamefont {M.}~\bibnamefont {Linkmann}}, \ and\
		\bibinfo {author} {\bibfnamefont {B.}~\bibnamefont {Eckhardt}},\ }\bibfield
	{title} {\enquote {\bibinfo {title} {Using machine learning to predict
				extreme events in the h{\'e}non map},}\ }\href@noop {} {\bibfield  {journal}
		{\bibinfo  {journal} {Chaos: An Interdisciplinary Journal of Nonlinear
				Science}\ }\textbf {\bibinfo {volume} {30}},\ \bibinfo {pages} {013113}
		(\bibinfo {year} {2020})}\BibitemShut {NoStop}%
	\bibitem [{\citenamefont {N{\"a}rhi}\ \emph {et~al.}(2018)\citenamefont
		{N{\"a}rhi}, \citenamefont {Salmela}, \citenamefont {Toivonen}, \citenamefont
		{Billet}, \citenamefont {Dudley},\ and\ \citenamefont {Genty}}]{narhi2018}%
	\BibitemOpen
	\bibfield  {author} {\bibinfo {author} {\bibfnamefont {M.}~\bibnamefont
			{N{\"a}rhi}}, \bibinfo {author} {\bibfnamefont {L.}~\bibnamefont {Salmela}},
		\bibinfo {author} {\bibfnamefont {J.}~\bibnamefont {Toivonen}}, \bibinfo
		{author} {\bibfnamefont {C.}~\bibnamefont {Billet}}, \bibinfo {author}
		{\bibfnamefont {J.~M.}\ \bibnamefont {Dudley}}, \ and\ \bibinfo {author}
		{\bibfnamefont {G.}~\bibnamefont {Genty}},\ }\bibfield  {title} {\enquote
		{\bibinfo {title} {Machine learning analysis of extreme events in optical
				fibre modulation instability},}\ }\href@noop {} {\bibfield  {journal}
		{\bibinfo  {journal} {Nature communications}\ }\textbf {\bibinfo {volume}
			{9}},\ \bibinfo {pages} {1--11} (\bibinfo {year} {2018})}\BibitemShut
	{NoStop}%
	\bibitem [{\citenamefont {Yeditha}\ \emph {et~al.}(2020)\citenamefont
		{Yeditha}, \citenamefont {Kasi}, \citenamefont {Rathinasamy},\ and\
		\citenamefont {Agarwal}}]{yeditha2020}%
	\BibitemOpen
	\bibfield  {author} {\bibinfo {author} {\bibfnamefont {P.~K.}\ \bibnamefont
			{Yeditha}}, \bibinfo {author} {\bibfnamefont {V.}~\bibnamefont {Kasi}},
		\bibinfo {author} {\bibfnamefont {M.}~\bibnamefont {Rathinasamy}}, \ and\
		\bibinfo {author} {\bibfnamefont {A.}~\bibnamefont {Agarwal}},\ }\bibfield
	{title} {\enquote {\bibinfo {title} {Forecasting of extreme flood events
				using different satellite precipitation products and wavelet-based machine
				learning methods},}\ }\href@noop {} {\bibfield  {journal} {\bibinfo
			{journal} {Chaos: An Interdisciplinary Journal of Nonlinear Science}\
		}\textbf {\bibinfo {volume} {30}},\ \bibinfo {pages} {063115} (\bibinfo
		{year} {2020})}\BibitemShut {NoStop}%
	\bibitem [{\citenamefont {Wan}\ \emph {et~al.}(2018)\citenamefont {Wan},
		\citenamefont {Vlachas}, \citenamefont {Koumoutsakos},\ and\ \citenamefont
		{Sapsis}}]{wan2018}%
	\BibitemOpen
	\bibfield  {author} {\bibinfo {author} {\bibfnamefont {Z.~Y.}\ \bibnamefont
			{Wan}}, \bibinfo {author} {\bibfnamefont {P.}~\bibnamefont {Vlachas}},
		\bibinfo {author} {\bibfnamefont {P.}~\bibnamefont {Koumoutsakos}}, \ and\
		\bibinfo {author} {\bibfnamefont {T.}~\bibnamefont {Sapsis}},\ }\bibfield
	{title} {\enquote {\bibinfo {title} {Data-assisted reduced-order modeling of
				extreme events in complex dynamical systems},}\ }\href {\doibase
		10.1371/journal.pone.0197704} {\bibfield  {journal} {\bibinfo  {journal}
			{PLOS ONE}\ }\textbf {\bibinfo {volume} {13}},\ \bibinfo {pages} {1--22}
		(\bibinfo {year} {2018})}\BibitemShut {NoStop}%
	\bibitem [{\citenamefont {Chattopadhyay}, \citenamefont {Nabizadeh},\ and\
		\citenamefont {Hassanzadeh}(2020)}]{hassanzadeh2020}%
	\BibitemOpen
	\bibfield  {author} {\bibinfo {author} {\bibfnamefont {A.}~\bibnamefont
			{Chattopadhyay}}, \bibinfo {author} {\bibfnamefont {E.}~\bibnamefont
			{Nabizadeh}}, \ and\ \bibinfo {author} {\bibfnamefont {P.}~\bibnamefont
			{Hassanzadeh}},\ }\bibfield  {title} {\enquote {\bibinfo {title} {Analog
				forecasting of extreme-causing weather patterns using deep learning},}\
	}\href {\doibase https://doi.org/10.1029/2019MS001958} {\bibfield  {journal}
		{\bibinfo  {journal} {Journal of Advances in Modeling Earth Systems}\
		}\textbf {\bibinfo {volume} {12}} (\bibinfo {year} {2020}),\
		https://doi.org/10.1029/2019MS001958}\BibitemShut {NoStop}%
	\bibitem [{\citenamefont {Rudy}\ and\ \citenamefont
		{Sapsis}(2021{\natexlab{b}})}]{Rudy2021b}%
	\BibitemOpen
	\bibfield  {author} {\bibinfo {author} {\bibfnamefont {S.~H.}\ \bibnamefont
			{Rudy}}\ and\ \bibinfo {author} {\bibfnamefont {T.~P.}\ \bibnamefont
			{Sapsis}},\ }\bibfield  {title} {\enquote {\bibinfo {title} {Prediction of
				intermittent fluctuations from surface pressure measurements on a turbulent
				airfoil},}\ }\href@noop {} {\bibfield  {journal} {\bibinfo  {journal} {AIAA}\
		} (\bibinfo {year} {2021}{\natexlab{b}})},\ \bibinfo {note} {in
		press}\BibitemShut {NoStop}%
	\bibitem [{\citenamefont {Chattopadhyay}, \citenamefont {Hassanzadeh},\ and\
		\citenamefont {Subramanian}(2020)}]{subramanian2019}%
	\BibitemOpen
	\bibfield  {author} {\bibinfo {author} {\bibfnamefont {A.}~\bibnamefont
			{Chattopadhyay}}, \bibinfo {author} {\bibfnamefont {P.}~\bibnamefont
			{Hassanzadeh}}, \ and\ \bibinfo {author} {\bibfnamefont {D.}~\bibnamefont
			{Subramanian}},\ }\bibfield  {title} {\enquote {\bibinfo {title} {Data-driven
				predictions of a multiscale lorenz 96 chaotic system using machine-learning
				methods: reservoir computing, artificial neural network, and long short-term
				memory network},}\ }\href {\doibase 10.5194/npg-27-373-2020} {\bibfield
		{journal} {\bibinfo  {journal} {Nonlinear Processes in Geophysics}\ }\textbf
		{\bibinfo {volume} {27}},\ \bibinfo {pages} {373--389} (\bibinfo {year}
		{2020})}\BibitemShut {NoStop}%
	\bibitem [{\citenamefont {Pyragas}\ and\ \citenamefont
		{Pyragas}(2020)}]{pyragas2020}%
	\BibitemOpen
	\bibfield  {author} {\bibinfo {author} {\bibfnamefont {V.}~\bibnamefont
			{Pyragas}}\ and\ \bibinfo {author} {\bibfnamefont {K.}~\bibnamefont
			{Pyragas}},\ }\bibfield  {title} {\enquote {\bibinfo {title} {Using reservoir
				computer to predict and prevent extreme events},}\ }\href {\doibase
		https://doi.org/10.1016/j.physleta.2020.126591} {\bibfield  {journal}
		{\bibinfo  {journal} {Physics Letters A}\ }\textbf {\bibinfo {volume}
			{384}},\ \bibinfo {pages} {126591} (\bibinfo {year} {2020})}\BibitemShut
	{NoStop}%
	\bibitem [{\citenamefont {Farazmand}\ and\ \citenamefont
		{Sapsis}(2016)}]{PRE2016}%
	\BibitemOpen
	\bibfield  {author} {\bibinfo {author} {\bibfnamefont {M.}~\bibnamefont
			{Farazmand}}\ and\ \bibinfo {author} {\bibfnamefont {T.~P.}\ \bibnamefont
			{Sapsis}},\ }\bibfield  {title} {\enquote {\bibinfo {title} {Dynamical
				indicators for the prediction of bursting phenomena in high-dimensional
				systems},}\ }\href {\doibase 10.1103/PhysRevE.94.032212} {\bibfield
		{journal} {\bibinfo  {journal} {Phys. Rev. E}\ }\textbf {\bibinfo {volume}
			{94}},\ \bibinfo {pages} {032212} (\bibinfo {year} {2016})}\BibitemShut
	{NoStop}%
	\bibitem [{\citenamefont {Blonigan}, \citenamefont {Farazmand},\ and\
		\citenamefont {Sapsis}(2019)}]{blonigan2019}%
	\BibitemOpen
	\bibfield  {author} {\bibinfo {author} {\bibfnamefont {P.~J.}\ \bibnamefont
			{Blonigan}}, \bibinfo {author} {\bibfnamefont {M.}~\bibnamefont {Farazmand}},
		\ and\ \bibinfo {author} {\bibfnamefont {T.~P.}\ \bibnamefont {Sapsis}},\
	}\bibfield  {title} {\enquote {\bibinfo {title} {Are extreme dissipation
				events predictable in turbulent fluid flows?}}\ }\href {\doibase
		10.1103/PhysRevFluids.4.044606} {\bibfield  {journal} {\bibinfo  {journal}
			{Phys. Rev. Fluids}\ }\textbf {\bibinfo {volume} {4}},\ \bibinfo {pages}
		{044606} (\bibinfo {year} {2019})}\BibitemShut {NoStop}%
	\bibitem [{\citenamefont {Ansmann}\ \emph {et~al.}(2013)\citenamefont
		{Ansmann}, \citenamefont {Karnatak}, \citenamefont {Lehnertz},\ and\
		\citenamefont {Feudel}}]{FHNparameters}%
	\BibitemOpen
	\bibfield  {author} {\bibinfo {author} {\bibfnamefont {G.}~\bibnamefont
			{Ansmann}}, \bibinfo {author} {\bibfnamefont {R.}~\bibnamefont {Karnatak}},
		\bibinfo {author} {\bibfnamefont {K.}~\bibnamefont {Lehnertz}}, \ and\
		\bibinfo {author} {\bibfnamefont {U.}~\bibnamefont {Feudel}},\ }\bibfield
	{title} {\enquote {\bibinfo {title} {Extreme events in excitable systems and
				mechanisms of their generation},}\ }\href {\doibase
		10.1103/PhysRevE.88.052911} {\bibfield  {journal} {\bibinfo  {journal} {Phys.
				Rev. E}\ }\textbf {\bibinfo {volume} {88}},\ \bibinfo {pages} {052911}
		(\bibinfo {year} {2013})}\BibitemShut {NoStop}%
	\bibitem [{\citenamefont {Farazmand}(2016)}]{faraz_adjoint}%
	\BibitemOpen
	\bibfield  {author} {\bibinfo {author} {\bibfnamefont {M.}~\bibnamefont
			{Farazmand}},\ }\bibfield  {title} {\enquote {\bibinfo {title} {An
				adjoint-based approach for finding invariant solutions of {N}avier-{S}tokes
				equations},}\ }\href {\doibase 10.1017/jfm.2016.203} {\bibfield  {journal}
		{\bibinfo  {journal} {J. Fluid Mech.}\ }\textbf {\bibinfo {volume} {795}},\
		\bibinfo {pages} {278--312} (\bibinfo {year} {2016})}\BibitemShut {NoStop}%
	\bibitem [{\citenamefont {Farazmand}\ and\ \citenamefont
		{Sapsis}(2017{\natexlab{b}})}]{Farazmand2017}%
	\BibitemOpen
	\bibfield  {author} {\bibinfo {author} {\bibfnamefont {M.}~\bibnamefont
			{Farazmand}}\ and\ \bibinfo {author} {\bibfnamefont {T.~P.}\ \bibnamefont
			{Sapsis}},\ }\bibfield  {title} {\enquote {\bibinfo {title} {A variational
				approach to probing extreme events in turbulent dynamical systems},}\ }\href
	{\doibase 10.1126/sciadv.1701533} {\bibfield  {journal} {\bibinfo  {journal}
			{Science Advances}\ }\textbf {\bibinfo {volume} {3}} (\bibinfo {year}
		{2017}{\natexlab{b}}),\ 10.1126/sciadv.1701533}\BibitemShut {NoStop}%
	\bibitem [{\citenamefont {Mor{\'e}}(1978)}]{levenberg}%
	\BibitemOpen
	\bibfield  {author} {\bibinfo {author} {\bibfnamefont {J.~J.}\ \bibnamefont
			{Mor{\'e}}},\ }\bibfield  {title} {\enquote {\bibinfo {title} {The
				levenberg-marquardt algorithm: Implementation and theory},}\ }in\ \href@noop
	{} {\emph {\bibinfo {booktitle} {Numerical Analysis}}},\ \bibinfo {editor}
	{edited by\ \bibinfo {editor} {\bibfnamefont {G.~A.}\ \bibnamefont
			{Watson}}}\ (\bibinfo  {publisher} {Springer Berlin Heidelberg},\ \bibinfo
	{address} {Berlin, Heidelberg},\ \bibinfo {year} {1978})\ pp.\ \bibinfo
	{pages} {105--116}\BibitemShut {NoStop}%
	\bibitem [{\citenamefont {Takens}(1981)}]{takens}%
	\BibitemOpen
	\bibfield  {author} {\bibinfo {author} {\bibfnamefont {F.}~\bibnamefont
			{Takens}},\ }\bibfield  {title} {\enquote {\bibinfo {title} {Detecting
				strange attractors in turbulence},}\ }in\ \href {\doibase 10.1007/BFb0091924}
	{\emph {\bibinfo {booktitle} {Dynamical Systems and Turbulence, Warwick
				1980}}},\ \bibinfo {editor} {edited by\ \bibinfo {editor} {\bibfnamefont
			{D.}~\bibnamefont {Rand}}\ and\ \bibinfo {editor} {\bibfnamefont
			{L.}~\bibnamefont {Young}}}\ (\bibinfo  {publisher} {Springer Berlin
		Heidelberg},\ \bibinfo {address} {Berlin, Heidelberg},\ \bibinfo {year}
	{1981})\ pp.\ \bibinfo {pages} {366--381}\BibitemShut {NoStop}%
	\bibitem [{\citenamefont {Hochreiter}\ and\ \citenamefont
		{Schmidhuber}(1997)}]{Schmidhuber1997}%
	\BibitemOpen
	\bibfield  {author} {\bibinfo {author} {\bibfnamefont {S.}~\bibnamefont
			{Hochreiter}}\ and\ \bibinfo {author} {\bibfnamefont {J.}~\bibnamefont
			{Schmidhuber}},\ }\bibfield  {title} {\enquote {\bibinfo {title} {{Long
					Short-Term Memory}},}\ }\href {\doibase 10.1162/neco.1997.9.8.1735}
	{\bibfield  {journal} {\bibinfo  {journal} {Neural Computation}\ }\textbf
		{\bibinfo {volume} {9}},\ \bibinfo {pages} {1735--1780} (\bibinfo {year}
		{1997})},\ \Eprint
	{http://arxiv.org/abs/https://direct.mit.edu/neco/article-pdf/9/8/1735/813796/neco.1997.9.8.1735.pdf}
	{https://direct.mit.edu/neco/article-pdf/9/8/1735/813796/neco.1997.9.8.1735.pdf}
	\BibitemShut {NoStop}%
	\bibitem [{\citenamefont {Sherstinsky}(2020)}]{Sherstinsky2018}%
	\BibitemOpen
	\bibfield  {author} {\bibinfo {author} {\bibfnamefont {A.}~\bibnamefont
			{Sherstinsky}},\ }\bibfield  {title} {\enquote {\bibinfo {title}
			{Fundamentals of recurrent neural network (rnn) and long short-term memory
				(lstm) network},}\ }\href {\doibase
		https://doi.org/10.1016/j.physd.2019.132306} {\bibfield  {journal} {\bibinfo
			{journal} {Physica D: Nonlinear Phenomena}\ }\textbf {\bibinfo {volume}
			{404}},\ \bibinfo {pages} {132306} (\bibinfo {year} {2020})}\BibitemShut
	{NoStop}%
	\bibitem [{\citenamefont {Schrauwen}, \citenamefont {Verstraeten},\ and\
		\citenamefont {Van~Campenhout}(2007)}]{schrauwen2007}%
	\BibitemOpen
	\bibfield  {author} {\bibinfo {author} {\bibfnamefont {B.}~\bibnamefont
			{Schrauwen}}, \bibinfo {author} {\bibfnamefont {D.}~\bibnamefont
			{Verstraeten}}, \ and\ \bibinfo {author} {\bibfnamefont {J.}~\bibnamefont
			{Van~Campenhout}},\ }\bibfield  {title} {\enquote {\bibinfo {title} {An
				overview of reservoir computing: theory, applications and implementations},}\
	}in\ \href@noop {} {\emph {\bibinfo {booktitle} {Proceedings of the 15th
				european symposium on artificial neural networks. p. 471-482 2007}}}\
	(\bibinfo {year} {2007})\ pp.\ \bibinfo {pages} {471--482}\BibitemShut
	{NoStop}%
	\bibitem [{\citenamefont {Zimmermann}(2021)}]{easyesn}%
	\BibitemOpen
	\bibfield  {author} {\bibinfo {author} {\bibfnamefont {R.}~\bibnamefont
			{Zimmermann}},\ }\href@noop {} {\enquote {\bibinfo {title} {Easyesn: a
				library for recurrent neural networks using echo state networks},}\ }\bibinfo
	{howpublished} {\url{https://github.com/kalekiu/easyesn/}} (\bibinfo {year}
	{2021}),\ \bibinfo {note} {version 0.1.6.1}\BibitemShut {NoStop}%
	\bibitem [{\citenamefont {Guth}\ and\ \citenamefont
		{Sapsis}(2019)}]{sapsis2019}%
	\BibitemOpen
	\bibfield  {author} {\bibinfo {author} {\bibfnamefont {S.}~\bibnamefont
			{Guth}}\ and\ \bibinfo {author} {\bibfnamefont {T.~P.}\ \bibnamefont
			{Sapsis}},\ }\bibfield  {title} {\enquote {\bibinfo {title} {Machine learning
				predictors of extreme events occurring in complex dynamical systems},}\
	}\href {\doibase 10.3390/e21100925} {\bibfield  {journal} {\bibinfo
			{journal} {Entropy}\ }\textbf {\bibinfo {volume} {21}} (\bibinfo {year}
		{2019}),\ 10.3390/e21100925}\BibitemShut {NoStop}%
	\bibitem [{\citenamefont {Davis}\ and\ \citenamefont
		{Goadrich}(2006)}]{davis2006}%
	\BibitemOpen
	\bibfield  {author} {\bibinfo {author} {\bibfnamefont {J.}~\bibnamefont
			{Davis}}\ and\ \bibinfo {author} {\bibfnamefont {M.}~\bibnamefont
			{Goadrich}},\ }\bibfield  {title} {\enquote {\bibinfo {title} {The
				relationship between precision-recall and {ROC} curves},}\ }in\ \href
	{\doibase 10.1145/1143844.1143874} {\emph {\bibinfo {booktitle} {Proceedings
				of the 23rd International Conference on Machine Learning}}},\ \bibinfo
	{series and number} {ICML '06}\ (\bibinfo  {publisher} {Association for
		Computing Machinery},\ \bibinfo {address} {New York, NY, USA},\ \bibinfo
	{year} {2006})\ p.\ \bibinfo {pages} {233–240}\BibitemShut {NoStop}%
	\bibitem [{\citenamefont {Saito}\ and\ \citenamefont
		{Rehmsmeier}(2015)}]{saito2015}%
	\BibitemOpen
	\bibfield  {author} {\bibinfo {author} {\bibfnamefont {T.}~\bibnamefont
			{Saito}}\ and\ \bibinfo {author} {\bibfnamefont {M.}~\bibnamefont
			{Rehmsmeier}},\ }\bibfield  {title} {\enquote {\bibinfo {title} {The
				precision-recall plot is more informative than the {ROC} plot when evaluating
				binary classifiers on imbalanced datasets},}\ }\href {\doibase
		10.1371/journal.pone.0118432} {\bibfield  {journal} {\bibinfo  {journal}
			{PLOS ONE}\ }\textbf {\bibinfo {volume} {10}},\ \bibinfo {pages} {1--21}
		(\bibinfo {year} {2015})}\BibitemShut {NoStop}%
	\bibitem [{\citenamefont {An}(1996)}]{trainingnoiseref2}%
	\BibitemOpen
	\bibfield  {author} {\bibinfo {author} {\bibfnamefont {G.}~\bibnamefont
			{An}},\ }\bibfield  {title} {\enquote {\bibinfo {title} {The effects of
				adding noise during backpropagation training on a generalization
				performance},}\ }\href {\doibase 10.1162/neco.1996.8.3.643} {\bibfield
		{journal} {\bibinfo  {journal} {Neural Computation}\ }\textbf {\bibinfo
			{volume} {8}},\ \bibinfo {pages} {643--674} (\bibinfo {year}
		{1996})}\BibitemShut {NoStop}%
	\bibitem [{\citenamefont {Zur}\ \emph {et~al.}(2009)\citenamefont {Zur},
		\citenamefont {Jiang}, \citenamefont {Pesce},\ and\ \citenamefont
		{Drukker}}]{trainingnoiseref}%
	\BibitemOpen
	\bibfield  {author} {\bibinfo {author} {\bibfnamefont {R.~M.}\ \bibnamefont
			{Zur}}, \bibinfo {author} {\bibfnamefont {Y.}~\bibnamefont {Jiang}}, \bibinfo
		{author} {\bibfnamefont {L.~L.}\ \bibnamefont {Pesce}}, \ and\ \bibinfo
		{author} {\bibfnamefont {K.}~\bibnamefont {Drukker}},\ }\bibfield  {title}
	{\enquote {\bibinfo {title} {Noise injection for training artificial neural
				networks: A comparison with weight decay and early stopping},}\ }\href
	{\doibase https://doi.org/10.1118/1.3213517} {\bibfield  {journal} {\bibinfo
			{journal} {Medical Physics}\ }\textbf {\bibinfo {volume} {36}},\ \bibinfo
		{pages} {4810--4818} (\bibinfo {year} {2009})},\ \Eprint
	{http://arxiv.org/abs/https://aapm.onlinelibrary.wiley.com/doi/pdf/10.1118/1.3213517}
	{https://aapm.onlinelibrary.wiley.com/doi/pdf/10.1118/1.3213517} \BibitemShut
	{NoStop}%
	\bibitem [{\citenamefont {Young}\ \emph {et~al.}(2015)\citenamefont {Young},
		\citenamefont {Rose}, \citenamefont {Karnowski}, \citenamefont {Lim},\ and\
		\citenamefont {Patton}}]{Young2015}%
	\BibitemOpen
	\bibfield  {author} {\bibinfo {author} {\bibfnamefont {S.~R.}\ \bibnamefont
			{Young}}, \bibinfo {author} {\bibfnamefont {D.~C.}\ \bibnamefont {Rose}},
		\bibinfo {author} {\bibfnamefont {T.~P.}\ \bibnamefont {Karnowski}}, \bibinfo
		{author} {\bibfnamefont {S.-H.}\ \bibnamefont {Lim}}, \ and\ \bibinfo
		{author} {\bibfnamefont {R.~M.}\ \bibnamefont {Patton}},\ }\bibfield  {title}
	{\enquote {\bibinfo {title} {Optimizing deep learning hyper-parameters
				through an evolutionary algorithm},}\ }in\ \href {\doibase
		10.1145/2834892.2834896} {\emph {\bibinfo {booktitle} {Proceedings of the
				Workshop on Machine Learning in High-Performance Computing Environments}}},\
	\bibinfo {series and number} {MLHPC '15}\ (\bibinfo {year}
	{2015})\BibitemShut {NoStop}%
	\bibitem [{\citenamefont {Karnatak}\ \emph {et~al.}(2014)\citenamefont
		{Karnatak}, \citenamefont {Ansmann}, \citenamefont {Feudel},\ and\
		\citenamefont {Lehnertz}}]{feudel2014}%
	\BibitemOpen
	\bibfield  {author} {\bibinfo {author} {\bibfnamefont {R.}~\bibnamefont
			{Karnatak}}, \bibinfo {author} {\bibfnamefont {G.}~\bibnamefont {Ansmann}},
		\bibinfo {author} {\bibfnamefont {U.}~\bibnamefont {Feudel}}, \ and\ \bibinfo
		{author} {\bibfnamefont {K.}~\bibnamefont {Lehnertz}},\ }\bibfield  {title}
	{\enquote {\bibinfo {title} {Route to extreme events in excitable systems},}\
	}\href {\doibase 10.1103/PhysRevE.90.022917} {\bibfield  {journal} {\bibinfo
			{journal} {Phys. Rev. E}\ }\textbf {\bibinfo {volume} {90}},\ \bibinfo
		{pages} {022917} (\bibinfo {year} {2014})}\BibitemShut {NoStop}%
\end{thebibliography}

\end{document}